\documentclass[twoside]{article}

\usepackage[preprint]{aistats2027}
\usepackage{amsmath,amssymb,amsthm,mathtools}
\usepackage{dsfont}
\usepackage{enumitem}
\usepackage{booktabs,array}
\usepackage{xspace}
\usepackage{xcolor}
\usepackage[round]{natbib}

\usepackage[colorlinks=true,linkcolor=blue!60!black,citecolor=green!40!black,urlcolor=blue!60!black]{hyperref}
\usepackage{cleveref}

\usepackage[ruled]{algorithm}   %
\usepackage{algpseudocode}

\theoremstyle{plain}
\newtheorem{theorem}{Theorem}[section]
\newtheorem{proposition}[theorem]{Proposition}
\newtheorem{lemma}[theorem]{Lemma}
\newtheorem{corollary}[theorem]{Corollary}
\theoremstyle{definition}
\newtheorem{definition}[theorem]{Definition}
\newtheorem{assumption}[theorem]{Assumption}

\theoremstyle{remark}

\crefname{lemma}{Lemma}{Lemmas}
\crefname{theorem}{Theorem}{Theorems}
\crefname{proposition}{Proposition}{Propositions}
\crefname{corollary}{Corollary}{Corollaries}
\crefname{remark}{Remark}{Remarks}
\crefname{definition}{Definition}{Definitions}
\crefname{assumption}{Assumption}{Assumptions}
\crefname{algorithm}{Algorithm}{Algorithms}
\crefname{section}{Section}{Sections}
\crefname{subsection}{Section}{Sections}
\crefname{subsubsection}{Section}{Sections}
\crefname{appendix}{Appendix}{Appendices}
\crefname{equation}{}{}
\creflabelformat{equation}{(#2#1#3)}

\newcommand{\R}{\mathbb{R}}
\newcommand{\E}{\mathbb{E}}
\newcommand{\Prob}{\mathbb{P}}
\newcommand{\ip}[2]{\langle #1,#2\rangle}
\newcommand{\norm}[1]{\lVert #1\rVert}
\newcommand{\psum}{\mathbin{\|}}

\newcommand{\LMO}{\mathrm{LMO}}
\newcommand{\Reg}{\mathrm{Reg}}
\newcommand{\eps}{\varepsilon}
\newcommand{\dir}{\mathrm{dir}}
\newcommand{\indVec}{\mathds{1}}
\newcommand{\Proj}{\Pi}
\newcommand{\step}[1]{\par\smallskip\noindent\textit{#1}\enspace}
\newcommand{\firststep}[1]{\textit{#1}\enspace}
\newcommand{\OPT}{\citet{WGKT25}\xspace}
\DeclareMathOperator{\conv}{conv}
\DeclareMathOperator{\spn}{span}
\DeclareMathOperator{\aff}{aff}
\DeclareMathOperator{\diam}{diam}
\DeclareMathOperator{\dist}{dist}
\DeclareMathOperator*{\argmin}{arg\,min}

\usepackage[colorinlistoftodos,backgroundcolor=black!5!white,bordercolor=red]{todonotes}

\begin{document}

\runningtitle{An $\Omega(T^{3/4})$ lower bound for projection-free online learning with linear minimization oracles}

\twocolumn[

\aistatstitle{An $\Omega(T^{3/4})$ lower bound for projection-free online learning\\ with linear minimization oracles}

\aistatsauthor{ Julien Weibel \And Pierre Gaillard }

\aistatsaddress{ 
  Inria, D.I. École Normale Supérieure, CNRS\\ PSL Research University\\
  \texttt{julien.weibel@inria.fr} 
  \And 
    Inria, D.I. École Normale Supérieure, CNRS\\ PSL Research University\\
  \texttt{pierre.gaillard@inria.fr}
  } 
 
  \aistatsauthor{  Wouter M. Koolen \And Adrien Taylor} 

  \aistatsaddress{ 
  Centrum Wiskunde \& Informatica, Amsterdam\\
  and Twente University, Enschede\\
  \texttt{wmkoolen@cwi.nl}
  \And 
    Université Grenoble Alpes, Inria, CNRS\\ Grenoble INP, LJK\\
  \texttt{adrien.taylor@inria.fr}
  } 
 
\runningauthor{Julien Weibel, Pierre Gaillard, Wouter M. Koolen, Adrien Taylor}

  ]

\begin{abstract}
This work considers projection-free online learning algorithms with linear minimization oracles for handling the decision set.
For Online Frank--Wolfe (OFW), regret upper bounds were already known for specific OFW tunings
[Hazan and Kale, 2012, Hazan, 2016]
which scaled as $T^{3/4}$ (where $T$ is the horizon length).
However, it remained an open question whether the regret of all OFW tunings scaled as $T^{3/4}$.

We answer that question, proving a universal worst-case regret lower bound $\Omega(T^{3/4} m^{-1/4})$ for any projection-free algorithm that makes at most $m$ linear minimization oracle calls per round. 
The analysis suggests an extended  OFW to resort to multiple linear optimizations per round,
for which we prove a regret upper bound $O(\max \{ T^{3/4} m^{-1/4}, \sqrt{T} \})$.
\end{abstract}

\section{Introduction}\label{sec:intro}

\label{sec:problem}

We consider online convex optimization (OCO) over a convex compact decision set $K\subset\R^n$ 
(also called, for brevity, a body) 
of diameter at most~$D$, where a learner and an adversary play for $T$ rounds. At round~$t$, the learner plays $x_t\in K$, the adversary reveals a convex $L$-Lipschitz loss $f_t$, and the learner suffers $f_t(x_t)$ and observes a subgradient $g_t \in \partial f_t(x_t)$. The quality of the sequence $x_0,x_1,\dots,x_{T-1}$ is then quantified by its \emph{regret} against the best fixed point in hindsight,
\[
\Reg_T=\sum_{t<T}f_t(x_t)-\min_{u\in K}\sum_{t<T}f_t(u).
\]
OCO is detailed in~\citep{hazanIntroductionOnlineConvex2016,orabona2019modern}.

Online gradient descent \citep{zinkevich2003online} guarantees the minimax rate $\Reg_T=O(LD\sqrt T)$ but requires a Euclidean projection onto $K$ at every round. For many sets of interest, such as nuclear-norm balls and flow, matching or matroid polytopes, the projection is the bottleneck, while a linear minimization oracle ($\LMO$) over $K$ is cheap, e.g., via a leading singular pair or a combinatorial subroutine. \emph{Projection-free} methods replace the projection by an $\LMO$, which returns a minimizer of $\ip d\cdot$ over $K$ for a given direction $d$. In offline convex optimization, this yields the classical Frank--Wolfe method~\citep{frank1956algorithm}. The Online Frank--Wolfe (OFW), introduced by~\citet{hazan2012projection}, calls the $\LMO$ once per round using the gradient of a regularized cumulative loss, and moves a fraction of the way towards the returned vertex. OFW guarantees $\Reg_T=O(LD\,T^{3/4})$ for convex Lipschitz losses \citep{hazan2012projection,hazanIntroductionOnlineConvex2016}.

The central question of this paper is whether the gap between the $O(\sqrt T)$ and the $O(T^{3/4})$ can be bridged. In particular, is the $O(T^{3/4})$ precisely the price to pay for accessing $K$ only through $m=O(1)$ linear optimization calls per round, or is it an artifact of the current analysis and parameter choices in OFW? The question is stated as open by \citet{Mha24} and by \OPT. The latter reports an optimized variant of the OFW obtained through performance estimation problems~\citep{goujaudPEPitComputerassistedWorstcase2024}, along with strong numerical evidence that no method within the class of OFW, that performs no more than $m$ $\LMO$ calls per round beats $O(T^{3/4})$. This work answers positively: in high dimension ($n\geq T$), $\Omega(T^{3/4} m^{-1/4})$ is the price of the $\LMO$ model.

\subsection{Related work}\label{sec:related}

In projection-free online learning, beating the $O(T^{3/4})$ rate with $m=1$ requires either stronger oracles than LMOs or extra assumptions on the decision set $K$ or on the losses. 
Membership oracles \citep{levy2019projection,mhammedi2022efficient} and separation oracles \citep{GK22,Mha24} lead to $\sqrt T$ rates. Here, we assume that the learner only has access to an $\LMO$. With one $\LMO$ call per round, strongly convex decision sets allow $T^{2/3}$ for convex losses \citep{Wan_Zhang_2021}, and curvature of the set can be exploited further \citep{Mha22b}. On polytopes, local $\LMO$s give $\sqrt T$ with constants depending on the dimension and geometry \citep{GH16}.

For decision sets of finite dimension $n$, Follow-the-Perturbed-Leader (FTPL)  achieves $O(n^{1/4}\sqrt{T})$ \citep{kalai2005efficient,ALST14} and $O(\sqrt{n} T^{2/3})$ \citep{hazan2020faster} on linear and smooth losses respectively, but with dimension-dependent multiplicative constants that blow up with~$n$. In particular, FTPL may achieve~$o(T^{3/4})$ regret when $n = o(T)$. So no lower bound of order $T^{3/4}$ can hold for randomized learners on bodies of dimension~$o(T)$, and the dimension of a hard instance must grow with $T$. In fixed dimension, polynomially many oracle calls also simulate a separation oracle by cutting-plane methods. Our lower bounds are therefore valid under a classical \emph{large-scale assumption}: 
the dimension must be larger than $2mT$ for span methods
(see definition in \Cref{sec:results})
and $\gtrsim(mT)^{7/2}$ for arbitrary learners.

Our lower-bound hard instance for span methods adapts the chain principle of the classical lower bounds for first-order methods restricted to the span of past gradients \citep{NY83,nesterov2018lectures}: the oracle reveals the instance one element at a time, so that after $k$ queries a span method knows nothing beyond the $k$-th element while the comparator sits at the end of the chain.
Our extension to all learners follows \citet{WS16} and \citet{CDHS20}: a random rotation hides the chain in a high-dimensional space, so that, with high probability, no query of the learner has a significant component along the directions not yet revealed, and the oracle answers as it would to a span method.

\subsection{Setting}\label{sec:setting}

We consider the following online learning framework: a \emph{learner} plays against an adversary over $T$ rounds, and the goal of the learner is to minimize its regret. At the start of the game, the adversary chooses an \emph{instance} of the game. 

\paragraph{Instance, oracle and learner.} An instance of the game is given by a triple $(K,\LMO,(g_t)_{t<T})$, fixed before the game (oblivious adversary), where:
\begin{itemize}[nosep, topsep=-\parskip, label={--},leftmargin=*]
    \item $K$ is a convex compact decision set in $\R^n$;
    \item $\LMO:\R^n\to K$ is a linear minimization oracle, that solves $\LMO(d)\in\argmin_{x\in K}\ip dx$. When the minimizer is not unique, the oracle may return any minimizer, and our instances fix a tie-breaking rule that the learner may know. 
    \item $(g_t)_{t<T}$ is a sequence of loss vectors in $\R^n$.
\end{itemize}
The starting point $x_0\in K$ is part of the instance, hence chosen by the adversary. 
Since the problem is invariant under translations, we fix $x_0=0$.

In each round $t=0,\dots,T-1$, the learner:
\begin{enumerate}[leftmargin=*,nosep,topsep=-\parskip]
\item makes (if $t\ge1$) at most $m$ oracle calls, taking the form of: 
for $1\le k\le m$,
it chooses a query $d_{t,k}\in\R^n$ and receives $\LMO(d_{t,k})\in K$,
\item plays a point $x_t$ that is certified to lie in $K$,
\item observes the loss vector $g_t$.
\end{enumerate}

\medskip
The goal of the learner is to minimize its regret
\begin{equation} \label{eq:regret}
\Reg_T=\sum_{t<T}\ip{g_t}{x_t}-\min_{u\in K}\sum_{t<T}\ip{g_t}u.
\end{equation}

It is worth emphasizing that the learner cannot project onto $K$. 
We nevertheless make the learner slightly more powerful by granting it two extra abilities: in our lower bound
(recall we fixed $x_0=0$) it knows a radius $r>0$ such that $B_r=\{x\in\R^n:\norm{x}\le r\}\subseteq K$, and it may use internal randomness $\xi$, independent of the instance. Queries and plays may be arbitrary measurable functions of $\xi$ and of everything observed so far. A point is \emph{certified} if it lies in $K$ for every instance consistent with information collected so far: for example, any point of the convex hull of $B_r$ and of the oracle's answers so~far is certified.

\paragraph{Why linear losses?} A linear loss $x\mapsto\ip{g_t}x$ with $\norm{g_t}\le L$ is convex and $L$-Lipschitz, and observing $g_t$ is equivalent to observing the whole loss $f_t$. So, a lower bound proved with linear losses and full information also holds for convex $L$-Lipschitz losses under every feedback model that includes a subgradient at the played point. Hence, we assume linear losses.

\subsection{Contributions}\label{sec:results}

\paragraph{Lower bound for span methods (\Cref{sec:span}).} Our first contribution (\Cref{thm:span}) is a single fixed instance, $(K,\LMO,(g_t)_{t<T})$, that may be known beforehand by the learner, and that is hard for a large subclass of learners called \emph{span methods}. This subclass restrict the learner to query $\LMO$s in the span of past observations and play points $x_t$ in the convex hull of the starting point and the past $\LMO$s' outputs (see \cref{def:span}). 
In particular, this subclass contains all OFW tunings considered by~\citet{WGKT25}, thereby answering their open question.

More precisely, the hard instance is defined by a set $K$ of dimension $2mT$ with diameter 1, and gradients $g_t$ with norms bounded by 1, such that any span method suffers a regret \[\Reg_T \geq T^{3/4} (4m)^{-1/4} .\] We restrict only the span of its queries and plays, which excludes randomized algorithms such as FTPL. 

\paragraph{Lower bound for all learners (\Cref{sec:all}).} Our second result and main contribution removes 
the span restriction on the learner,
at the price of a random instance and a larger dimension $n$. More precisely, Theorem~\ref{thm:all} states that there exists a fixed distribution of instances with $K$ of dimension $n\gtrsim(mT)^{7/2}$ and $L=D=1$ such that every learner that makes at most $m$ oracle calls per round satisfies 
\[
    \E[\Reg_T]\ \ge\ 0.23\,T^{3/4} m^{-1/4}\,,
\]
where the expectation is over both the instance and the learner's randomness.

\paragraph{Upper bound for OFW with $m$ LMO calls per round (\Cref{sec:ub}).}  \OPT gave numerical evidence that no OFW tuning making $m=O(1)$ LMO calls per round beats the $T^{3/4}$ regret, but left open how the regret depends on $m$. Our third result shows that the right dependence is $m^{-1/4}$. We introduce $m$-step OFW (\Cref{alg:mofw-opt}), which generalizes OFW to $m$ LMO calls per round, and prove that its regret matches the lower bound of \Cref{thm:span}  for $m\le T$ up to a multiplicative constant. Adapting the potential-based proof of \citet{WGKT25}, we show
\begin{equation}
    \label{eq:regret_upper-bound}
    \Reg_T \leq 4.8\, LD  \max \big\{ T^{3/4} m^{-1/4}, \sqrt{T} \big\} \,,
\end{equation}
for well-chosen hyperparameters. When $m = 1$, it recovers the rate of \citet{WGKT25}, while when $m \geq T$, it recovers the $O(\sqrt{T})$ minimax rate of OCO. 
We also show that~\eqref{eq:regret_upper-bound} holds,
with a larger constant, for an anytime tuning of m-step OFW, which does not need to know the horizon $T$.

\section{Lower bound for span methods}\label{sec:span}

In this section, we prove the lower bound for the specific subclass of learners called \emph{span methods}. The latter is large enough to contain OFW \citep{hazan2012projection} with any number of LMO calls per round, its variants with adaptive step sizes or line search, away-step and pairwise variants, and the generalized OFW tunings of \citet{WGKT25}, whose queries combine past gradients and vertices with arbitrary fixed coefficients. However, it excludes methods that compute projections, such as online gradient descent, or that randomly perturb their queries, such as~FTPL.

\begin{definition}[Span methods]\label{def:span}
A learner is a \emph{span method} if:
\begin{itemize}[leftmargin=0pt,
    labelsep=1em,
    labelwidth=.5em,
    itemindent=1.5em, topsep=0pt,itemsep=0pt,parsep=0pt,topsep=-\parskip]
\item its LMO queries $d_{t,k}$ are linear combination of the starting point $x_0$, past LMO answers $v_{s,j}$ and observed loss vectors $g_s$:  
    for all $t\ge 1$
    and $1\leq k \leq m$,
 \begin{align*}
    d_{t,k} & \in \spn\!\big( x_0, v_{1,1}, \dots, v_{1,m}, \dots, v_{t,k-1}, g_0, \dots, g_{t-1} \big)\\
    v_{t,k} & = \LMO(d_{t,k})\,;
    \end{align*}
\item {its play points $x_t$ are convex combination of past LMO answers:}
    for all rounds $t\ge 0$, 
\[x_t \in \conv\big( \{ x_0\} \cup \{ v_{1,1}, \dots, v_{1,m}, \dots, v_{t,m} \} \big)\,.
    \]
\end{itemize}
\end{definition}

Note that only the sets of vectors from which $d_{t,k}$ and $x_t$ are formed are restricted;
the linear and convex coefficients used to form $d_{t,k}$ and $x_t$  may be any functions of everything the learner knows, including its internal randomness, the decision set, the oracle's rule, or even future losses. 

\begin{theorem}[Lower bound for span methods]
\label{thm:span}
Let $m,T\ge1$ and $N=m(T-1)+1$. There is a fixed instance with $K \subset \R^{2N}$, $D = 1$, and $\norm{g_t}=1$, on which every span method making at most $m$ oracle calls per round has regret %
\[ \Reg_T\ge T\,(4N)^{-1/4} \ge T^{3/4}\, (4m)^{-1/4}.\]
\end{theorem}
The proof, deferred to Appendix~\ref{app:proof_thmspan}, is based on constructing a single instance that is hard for all span methods.
The hard-instance construction for our lower bound 
adapts the chain principle of the classical lower bounds for first-order methods restricted to the span of past gradients \citep{NY83,nesterov2018lectures}.
Scaling the instance, for all $L,D>0$ there is an instance with $\diam K\le D$ and $\norm{g_t}\le L$ on which every span method 
making at most $m$ oracle calls per round
has regret at least $LD\,T^{3/4}\,(4m)^{-1/4}$.

The rest of this section is devoted to providing the key ideas of the proof of Theorem~\ref{thm:span}.

\subsection{Design of the hard instance}\label{sec:span-instance}
We start by constructing the hard instance of~\Cref{thm:span}.
The body is a simplex whose vertices are the origin $x_0 = 0$ and $N=m(T-1)+1$ points called the \emph{atoms}. %
The atoms form a chain: they share a common walk with orthogonal steps, a deterministic analogue of a random walk, to which each atom adds a private orthogonal coordinate.
The losses reward all atoms beyond a moving index by the same amount. 
The oracle breaks ties in favor of earlier atoms to force the learner to walk the chain one atom at a time.

\paragraph{Body and its atoms.} 
Let $u_0,\nu_0,\dots,u_{N-1},\nu_{N-1}$ be the standard basis of $\R^{2N}$. The vectors $u_i$ carry the steps of the common walk, %
and $\nu_j$ carries the private coordinate of atom $\bar w_j$:
\begin{equation*}
\bar w_j=M_j+\sqrt{f_j}\,\nu_j,\qquad M_j=\sum_{i\le j}\sqrt{\delta_i}\,u_i,
\end{equation*}
where the \emph{increments} $\delta_j=e_j-e_{j-1}$ and \emph{noise sizes} $f_j=\min(e_j,1-e_j)$
depend on a \emph{profile} $0=e_{-1} < e_0 < e_1 < \dots < e_{N-1}$
which we define in the next paragraph.
The body is the following simplex
\begin{equation*}
\bar K=\conv\big(\{0\}\cup\{\bar w_0,\dots,\bar w_{N-1}\}\big).
\end{equation*}
The starting point imposed on the learner is $x_0=0$.
Any two atoms share the beginning of the walk, which gives the \emph{martingale identity} $\ip{\bar w_i}{\bar w_j}=e_i$ for $i<j$, and $\norm{\bar w_j}^2=e_j+f_j\le1$. Each atom is the only generating point with a component along its $\nu_j$, so the $N+1$ points are affinely independent and $\bar K$ is a simplex; its diameter is at most $1$ (\cref{lem:profile}(f)). 
The oracle $\LMO$ returns the first minimizer of $\ip d\cdot$ in the list $0,\bar w_0,\bar w_1,\dots,\bar w_{N-1}$.

\paragraph{The profile.} For $k\ge0$ let $h_k=\sqrt{2k+1}$ and $\hat h_k=h_k/(2\sqrt N)$. For $0\le j\le N-1$ the \emph{profile} $e_j$ is
\[
e_j=\begin{cases}\hat h_j,& j<N/2,\\ 1-\hat h_{N-1-j},& j\ge N/2.\end{cases}
\]
The profile rises like a square root from $0$ to $\frac12$ and then symmetrically up to $1$. It is strictly increasing (\cref{lem:profile}), and $\delta_jf_j$ is of order $1/N$ at every index, which is what makes every round cost roughly the same amount, of order $N^{-1/4}$ (\cref{sec:span-size}).

\paragraph{The losses.} Let $\bar W$ be the matrix with the atoms as columns, 
$A = \bar W^\top \bar W$ the Gram matrix of the atoms, 
and $\indVec_\kappa$ the indicator vector of the indices $j\ge\kappa$. The atoms are linearly independent, so $A$ is invertible, and for $0\le\kappa\le N-1$ we set
\[
\bar c_\kappa=\big(\indVec_\kappa^\top A^{-1}\indVec_\kappa\big)^{-1/2},\qquad
\bar g^{(\kappa)}=-\bar c_\kappa\,\bar WA^{-1}\indVec_\kappa .
\]

Then $\bar W^\top\bar g^{(\kappa)}=-\bar c_\kappa\indVec_\kappa$ and $\norm{\bar g^{(\kappa)}}^2=\bar c_\kappa^2\,\indVec_\kappa^\top A^{-1}\indVec_\kappa=1$: 
$\bar g^{(\kappa)}$ is the unit vector in the span of the atoms with
\begin{equation}\label{eq:canon}
\ip{\bar g^{(\kappa)}}{\bar w_j}=\begin{cases}0,& j<\kappa,\\[2pt] -\bar c_\kappa,& j\ge\kappa.\end{cases}
\end{equation}
It gives no reward to the atoms before $\kappa$, and it rewards every atom from $\kappa$ on by the same amount. The learner faces the fixed loss schedule
\[
g_t=\bar g^{(\kappa_t)},\qquad\kappa_t=mt,\qquad t=0,\dots,T-1,
\]
every $\kappa_t$ being a valid index because $m(T-1)=N-1$. The schedule is matched to the oracle budget: when it plays in round $t$, the learner has made at most $\kappa_t$ oracle calls. The comparator is the last atom $x^*=\bar w_{N-1}$ and is  rewarded in every round. Everything is deterministic and may be known to the learner.

\subsection{Every round costs \texorpdfstring{$\bar c_{\kappa_t}$}{c}}\label{sec:span-ties}

The key idea is that a query built from past losses and past atoms gives the same value to all atoms beyond the current index, so the oracle, which breaks ties in favor of the earliest atom, advances along the chain by at most one atom per call.

\begin{lemma}\label{lem:ties}
Let $t\ge 1$ and $J \ge \kappa_{t-1}=m(t-1)$.
Let $d\in\spn(0,\bar w_0, \dots, \bar w_{J-1},g_0,\dots, g_{t-1})$.
Then, for all $j \geq J$, we have $\ip{d}{\bar w_j} = \ip{d}{\bar w_J}$.

In particular, the oracle tie-breaking rule gives $\LMO(d) \in \{0, \bar w_0, \dots, \bar w_J\}$.
\end{lemma}%

\textit{Proof sketch. (The complete proof is in \cref{app:proof_ties}.)}
A received loss $g_s = \bar g^{(\kappa_s)}$ has $\kappa_s\le J$ and gives the value $-\bar c_{\kappa_s}$ to every atom of index at least $J$ by \eqref{eq:canon}, a returned atom $\bar w_i$ has $i<J$ and gives them the value $e_i$ by the martingale identity, and linear combinations preserve this.
Using \Cref{lem:ties}, we get that all span methods incur the same exact regret on this instance. %
\begin{proposition}\label{prop:span}
On the instance of \cref{sec:span-instance}, every span method making at most $m$ oracle calls per round has regret exactly $\Reg_T = \sum_{t<T}\bar c_{\kappa_t}$.
\end{proposition}

\textit{Proof sketch. (The complete proof is in \cref{app:proof_propspan}.)}
By induction on $t$ and \cref{lem:ties}, every atom returned before the play of round $t$ has index below $\kappa_t=mt$: round $0$ makes no call, the oracle advances by at most one atom per call, at most $m$ calls are made per round, and the schedule advances by $m$ atoms per round. So the play $x_t$ combines the origin and atoms of index below $\kappa_t$, which all have loss $0$ against $g_t$ by \eqref{eq:canon}, while the comparator has loss $-\bar c_{\kappa_t}$ and is the best point in hindsight. This is where $m$ enters: the schedule has to advance $m$ atoms per round, which costs the adversary a factor $m$ in the number of atoms and hence $m^{-1/4}$ in the regret.

\subsection{A lower bound on per-round costs \texorpdfstring{$\bar c_\kappa$}{c}}\label{sec:span-size}

We prove the following lower bound on $\bar c_\kappa$, which combined with \Cref{prop:span} gives \Cref{thm:span}.
\begin{lemma}%
\label{lem:cbar}
For $0\le\kappa\le N-1$, 
we have \[ \bar c_\kappa = \dist\big( \aff(\bar w_\kappa,\dots,\bar w_{N-1}),\ \spn(\bar w_0,\dots,\bar w_{\kappa-1})  \big) ,\]
and
\[
\bar c_\kappa\ \ge\ \dist\big(M_\kappa,\ \spn(\bar w_0,\dots,\bar w_{\kappa-1})\big)\ \ge\ (4N)^{-1/4}.
\]
\end{lemma}

\textit{Proof sketch. (The complete proof is in \cref{app:proof:cbar}.)}\\
\emph{The distance.} Let $y$ be in the affine hull of the tail atoms and $z$ in the span of the head atoms. By \eqref{eq:canon}, $\ip{\bar g^{(\kappa)}}{y-z}=-\bar c_\kappa$, so $\norm{y-z}\ge\bar c_\kappa$ because $\bar g^{(\kappa)}$ is a unit vector. Equality is attained: $\bar g^{(\kappa)}=\sum_j\beta_j\bar w_j$ with $\beta=-\bar c_\kappa A^{-1}\indVec_\kappa$, and $1=\norm{\bar g^{(\kappa)}}^2=\sum_j\beta_j\ip{\bar g^{(\kappa)}}{\bar w_j}=-\bar c_\kappa\sum_{j\ge\kappa}\beta_j$ by \eqref{eq:canon}, so $y^*=-\bar c_\kappa\sum_{j\ge\kappa}\beta_j\bar w_j$ is in the affine hull of the tail, $z^*=\bar c_\kappa\sum_{j<\kappa}\beta_j\bar w_j$ in the span of the head, and $y^*-z^*=-\bar c_\kappa\,\bar g^{(\kappa)}$ has norm $\bar c_\kappa$.\\
\emph{The first inequality.} Every tail atom is $M_\kappa$ plus a vector orthogonal to $M_\kappa$ and to the head atoms, so $\norm{y-z}^2=\norm{M_\kappa-z}^2+\norm{y-M_\kappa}^2$: the tail is never closer to the head span than $M_\kappa$ is.\\
\emph{The second inequality.} 
Expand $M_\kappa-\sum_{s<\kappa}y_s\bar w_s$ along the vectors $u_i$ and $\nu_s$ and set $Z_i=1-\sum_{s=i}^{\kappa-1}y_s$ (so that $Z_\kappa=1$): its squared norm is a discrete Dirichlet energy on the chain,
and $\dist(M_\kappa,\spn(\bar w_0,\dots,\bar w_{\kappa-1}))^2$ is the minimum of this energy,
\[
q_\kappa:=\min_{Z:\,Z_\kappa=1}\ \sum_{i\le\kappa}\delta_i\,Z_i^2+\sum_{s<\kappa}f_s\,(Z_s-Z_{s+1})^2 .
\]
Minimizing over $Z_0,Z_1,\dots$ one variable at a time, each step being $\min_t[x\,t^2+y\,(t-1)^2]=xy/(x+y)$
(the formula for two resistances in parallel), 
gives the recursion $q_0=\delta_0$, $q_{k+1}=\delta_{k+1}+q_kf_k/(q_k+f_k)$, and on our profile $\sqrt N\,q_k\ge\frac12$ for every $k$ (\cref{lem:qrec}). Only the head of the chain enters this bound.

This completes the proof of \Cref{thm:span}.

\subsection{Remarks}\label{sec:span-discussion}

\paragraph{Tie-breaking.} The instance uses the oracle's tie rule, because the tail ties exactly. 
The tie rule can be removed (up to ties with the origin) by
shrinking atom $j$ by $\rho_j = 1-\eps j/(N-1)$ (that is replacing $\bar w_j$ by $\tilde w_j = \rho_j \bar w_j$)
for a small $\eps>0$,
at the price of a factor $1-\eps$ in the regret bound.\\
Indeed, two shrunk atoms satisfy $\ip{\tilde w_i}{\tilde w_j}=\rho_i\rho_je_i$ for $i<j$, and $\ip{\bar g^{(\kappa)}}{\tilde w_j}=-\rho_j \bar c_\kappa$ for $j\ge\kappa$. So every query of a span method gives each tail atom $\rho_j$ times a common number. If the common number is negative, the first tail atom is the strict best of the tail; if it is positive, the whole tail loses to the origin. 
Otherwise the tail ties with the origin
(e.g., for the query~$0$), which the oracle breaks in favor of the origin. 

\paragraph{On the multiplicative constant $4^{-1/4}$.}
Comparing the lower bound of \Cref{thm:span} to the upper bounds of \OPT for $m=1$
and of Section~\ref{sec:ub} for $m>1$,
we get that the gap is a multiplicative factor $4\times 3^{-3/4}\, \sqrt{2} < 2.5$. 
Numerically, for $m=1$, $\sum_{t<T}\bar c_t$ exceeds $T^{3/4}$ at every horizon we computed (\Cref{tab:comparison}), so the constant $4^{-1/4}$ of \Cref{thm:span} is not sharp and the true constant appears to be around $1$.

\paragraph{The instance is hard only for span methods: FTPL beats it.} 
FTPL uses one oracle call per round but perturbs its query with a  full-dimensional random noise, so it is not a span method. 
On the convex hull of $N+1$ known points, FTPL with Gaussian perturbations has regret $O(LD\sqrt T)$ up to a factor logarithmic in $N$ \citep{ALST14}, 
and Table~\ref{tab:comparison} illustrates this numerically on the instance of Section~\ref{sec:span-instance}.
To design a lower bound for a larger class of learners that includes FTPL, the chain of atoms has to be hidden %
by a random rotation in a space of much larger dimension: this is the instance of \cref{sec:all}.

\begin{table}[h!]
\begin{center}\small
\begin{tabular}{@{}cccc@{}}
\toprule
$T$  & $T^{3/4}$ & Span methods ($\sum_t \bar c_t$) & FTPL\\
\midrule
16 & 8.000 & 8.066 & 2.46\\
32 &  13.454 & 13.528 & 3.18\\
64 &  22.627 & 22.705 & 4.30\\
128 &  38.055 & 38.133 & 6.29\\
256 &  64.000 & 64.077 & 10.90\\
\bottomrule
\end{tabular}
\end{center}
\caption{Regret lower bound for span methods with $m=1$ LMO call per round, together with regret of FTPL (with theoretical horizon-dependent tuning) on the hard instance of Section~\ref{sec:span-instance}.}
\label{tab:comparison}
\end{table}

\section{Lower bound for all learners}\label{sec:all}

A general learner may query the LMO in arbitrary (possibly random) directions, and may play outside the convex hull of the oracle's answers, by combining with elements of $B_r \subset K$. 

To handle such learners, we adapt the random rotation construction \citep{WS16,CDHS20}
to hide the orientation of the meaningful hard body of \Cref{sec:span}
in a uniformly random frame of a high-dimensional space, and add a shrinkage and a small ball.
This makes each oracle call reveal at most one new atom whatever the query.

\begin{theorem}[Lower bound for every learner]\label{thm:all}
Let $m\ge1$, $T\ge2$, $N=m(T-1)+1$ and $n\ge1.1\cdot10^{6}\,N^{7/2}$. There is a probability distribution over instances with $K\subset\R^n$, $0\in K$, $\diam K\le\sqrt5/2$ and $\norm{g_t}=1$, such that every learner making at most $m$ oracle calls per round satisfies
\[
\Reg_T\ \ge\ 0.25\,T\,N^{-1/4} 
\]
with probability at least $1-\frac1{100}N^{-1/4}$, and
\[
\E[\Reg_T]\ \ge\ 0.23\,T\,N^{-1/4},
\]
where the probability and the expectation are over the instance and the learner's randomness.
\end{theorem}

The instance adversary is still oblivious:
the instance distribution is the same for all learners,
and the instance randomness is independent of the learner's randomness (as a lower bound for randomized learners requires).
Because of this, by Yao's principle \citep{Yao77} it suffices to treat deterministic learners.

As a corollary of \Cref{thm:all}, choosing a learner-dependent rotation, 
we get a (deterministic) instance
that depends on the learner's distribution but not on its randomness: it is a lower bound in the usual sense for randomized algorithms.
\begin{corollary}[A fixed instance for each learner]
\label{cor:all}
Under the assumptions of \cref{thm:all}, for all $L,D>0$ and every learner making at most $m$ oracle calls per round, there is a fixed instance with $\diam K\le D$ and $\norm{g_t}\le L$ on which $\E_\xi[\Reg_T]\ge0.2\,LD\,T\,N^{-1/4}$.
\end{corollary}

The rest of the section is devoted to sketching the key ideas of the proof of Theorem~\ref{thm:all}, starting with the design of the hard hidden instance. The full proofs are deferred to Appendix~\ref{app:all}.

\subsection{The hidden instance}\label{sec:all-instance}

We start by constructing the hard instance of \Cref{thm:all}.
The instance is the chain of \cref{sec:span-instance} hidden by a random rotation in a space of dimension $n\gg N$, so that the learner cannot compute the atoms.
Two additions ensure that, with high probability, each oracle call reveals at most one new atom of the chain.
A small $n$-dimensional ball is added to the body, so that a query unrelated to the revealed atoms is answered by a ball point that reveals nothing.
The atoms are also slightly shrunk, so that a query with a tiny component on the unrevealed atoms 
can reveal only the first unrevealed atom.
\paragraph{Body and its atoms.} 
Let $n \gg 2N$ to be chosen later (see the end of Section~\ref{sec:all-outline}),
and let $(u_0,\nu_0,u_1,\nu_1,\dots,u_{N-1},\nu_{N-1})$ be a uniformly random orthonormal family in $\R^n$, called the \emph{frame} and listed in the order in which it will be revealed to the learner.
The atoms are those of \cref{sec:span-instance} in this frame, slightly shrunk:
\[
w_j=\rho_jM_j+\sqrt{f_j}\,\nu_j,\qquad\rho_j=1-\frac{j}{2(N-1)},
\]
where $M_j=\sum_{i\le j}\sqrt{\delta_i}\,u_i$ is the walk position in the hidden frame. The shrinkage factor decreases from $1$ to $\frac12$ by the same amount $\gamma=1/(2(N-1))$ at every step. Its only purpose is to break ties between atoms in favor of the earliest one. The body is
\[
K=\conv\big(\{0\}\cup\{w_0,\dots,w_{N-1}\}\cup B_r\big),
\]
with $B_r$ the closed ball of radius $r$ and $r=\tfrac1{40}N^{-1/4}$.
The starting point imposed on the learner is $x_0=0$.
For $d\ne0$ the oracle $\LMO$ returns the first candidate in the list $0,w_0,\dots,w_{N-1},-r\,d/\norm d$ that minimizes $\ip d\cdot$ (and $0$ for $d=0$). A linear function is minimized over a convex hull at a generating point and over the ball at $-r\,d/\norm d$, so this is a valid oracle. 

The ball plays two roles. For queries, it is a decoy: a query with no significant component along the revealed atoms is answered by the ball point $-r\,d/\norm d$, which reveals nothing (Step~1 of \cref{sec:all-outline}). For plays, it lets the learner leave the hull of the oracle's answers, but a ball point earns at most $r$ per round, a small fraction of what the comparator will earn. Finally, since the frame is random and $n\gg N$, the unrevealed atoms are nearly orthogonal to any vector the learner can form: a query cannot aim at them, and a play cannot put significant weight on them.

\paragraph{Losses.} The canonical loss $\bar g^{(\kappa)}$ of \cref{sec:span-instance} depends on all atoms through $A^{-1}$, so it cannot be computed from the part of the frame revealed up to index~$\kappa$. We use instead the unit vector pointing against the part of the walk position $M_\kappa$ that the first $\kappa$ atoms cannot explain: with $\Pi_\kappa$ the orthogonal projection onto $\spn(w_0,\dots,w_{\kappa-1})$,
\[
c_\kappa=\norm{M_\kappa-\Pi_\kappa M_\kappa},\qquad g^{(\kappa)}=-\frac{M_\kappa-\Pi_\kappa M_\kappa}{c_\kappa},
\]
and $g_t=g^{(\kappa_t)}$ with $\kappa_t=mt$ as before. 
Note that $c_\kappa>0$ since $M_\kappa\not\in\spn(w_0,\dots,w_{\kappa-1})$, and thus $g^{(\kappa)}$ is well defined.
This vector lies in $E_\kappa=\spn(u_0,\dots,u_\kappa,\nu_0,\dots,\nu_{\kappa-1})$ and has the same pattern as before up to the shrinkage (\cref{lem:F1}): $\ip{g^{(\kappa)}}{w_j}=0$ for $j<\kappa$ and $-\rho_jc_\kappa$ for $j\ge\kappa$, while every ball point has value at least $-r$. The comparator $x^*=w_{N-1}$ earns exactly $c_\kappa/2$ per round, and $c_\kappa\ge(4N)^{-1/4}$ 
by a recursion similar to that of \cref{lem:cbar} (the shrinkage only helping, \cref{lem:level}).

\subsection{Proof outline}\label{sec:all-outline}

As noted after \Cref{thm:all}, the instance is oblivious, and we can assume without loss of generality that the learner is deterministic. Below, we sketch the main ingredients of the proof of \Cref{thm:all}; all steps are formalized as lemmas in Appendix~\ref{app:all}.

In round $t$ only atoms of index at least $\kappa_t$ are useful, each with loss of order $-N^{-1/4}$, the comparator included. A learner that knows only atoms of index below $\kappa_t$ loses about $N^{-1/4}$ against $x^*$ in round $t$, as in \cref{sec:span}.

\paragraph{Step 1. We define a clean oracle and a clean process that reveal atoms one at a time.} 
The process is at \emph{level} $L$ when the first $2L+1$ frame vectors $u_0,\nu_0,\dots,\nu_{L-1},u_L$ are revealed. 
The atoms $w_0,\dots,w_{L-1}$ and the walk position $M_L$ are then functions of revealed vectors, and every unrevealed atom splits into a known part and a hidden part,
\[
w_j=\rho_jM_L+a^{(L)}_j,\qquad a^{(L)}_j\in U_L,
\]
where $U_L:=\spn(\nu_L,u_{L+1},\nu_{L+1},\dots,\nu_{N-1})$ is the span of the unrevealed frame vectors and $\norm{a^{(L)}_j}\le1$. To the learner, unrevealed atoms look alike except for the shrinkage. Before its play in round $t$ it has made at most $\kappa_t$ calls, so it can know an atom of index $\kappa_t$ or more only if some past oracle answer skipped an atom in the chain.
Two features prevent this:
\begin{itemize}[nosep, label={-}, leftmargin=*, topsep=-\parskip]
    \item \emph{Shrinkage:} for a query $d$ nearly orthogonal to $U_L$, every unrevealed atom has value $\rho_j\ip d{M_L}$ up to a tiny error, and if this is negative the next atom $w_L$ is the strict winner among unrevealed atoms by a margin $\gamma|\ip d{M_L}|$. 
    \item \emph{The ball:} an unrevealed atom can beat the ball point only if $|\ip d{M_L}|\gtrsim r\norm d$, and then the shrinkage margin $\gamma r\norm d$ exceeds the error from the hidden parts. %
    Otherwise no unrevealed atom is returned, and the oracle answer (the origin, a revealed atom, or $-r\,d/\norm d$) tells the learner nothing new.
\end{itemize}
A \emph{clean oracle} that answers using only the revealed vectors and raises the level by at most one per query (\cref{def:clean}) formalizes this. It agrees with the true oracle on every query $d$ such that $\norm{\Proj_{U_L}d}<\tau\norm d$, where $\Proj_{U_L}$ is the orthogonal projection onto $U_L$ and $\tau=\gamma r/5$ (\cref{lem:agree}). The \emph{clean process} is the run of the learner against the clean oracle: when the learner plays in round~$t$, its level is at most $\kappa_t$ (\cref{lem:count}).

\paragraph{Step 2. The clean and true processes coincide (with high probability).} In the clean process, where the true oracle is replaced by the clean oracle, every message of the learner, query or play, is a function of the revealed vectors (\cref{lem:prefix}). The messages are thus determined by the revealed vectors, but they can be outside their span. 

Conditionally on them, the unrevealed frame is a uniformly random orthonormal family of the orthogonal complement, so $U_L$ is a uniformly random subspace of dimension at most $2N$ in a space of dimension at least $n-2N$, and a fixed vector has projection of order $\sqrt{N/n}$ onto it (\cref{lem:proj}). A union bound over the at most $2N$ messages shows that, with probability at least $1-\delta$, where $\delta:=\frac1{100}N^{-1/4}$, every message $v$ satisfies $\norm{\Proj_{U_L}v}<\tau\norm v$ (\cref{prop:good}). We call this event the \emph{good event}. On this good event the learner's process is the clean process.

\paragraph{Step 3. On the good event, each certifiable play for round $t$ earns at most $0.11\,c_{\kappa_t}+r$, against $c_{\kappa_t}/2$ for the comparator.}
The learner may play any point of $K$ it can certify, for instance a mixture of unrevealed atoms and a ball point whose hidden parts cancel. This fails because the hidden parts of unrevealed atoms cannot cancel each other: a nonnegative combination of them with total weight $\Lambda$ has norm at least $\Lambda/(4N^{1/4})$ (\cref{lem:cancel}). %
A play with hidden component at most $\tau$ therefore puts weight at most $4N^{1/4}(\tau+r)\le0.11$ on unrevealed atoms and earns at most $0.11\,c_{\kappa_t}+r$ (\cref{lem:earn}), against $c_{\kappa_t}/2$ for the comparator.

\paragraph{Conclusion.} On the good event each round costs at least $0.39\,c_{\kappa_t}-r\ge0.25\,N^{-1/4}$.
And outside the good event, the regret is at least $-DT \ge -1.12\,T$.
As $\delta=\frac1{100}N^{-1/4}$, this gives $\E[\Reg_T]\ge0.23\,TN^{-1/4}$
and proves \Cref{thm:all}.
The dimension $n$ comes from the condition $\sqrt{N/n}\lesssim\tau$: the typical size of the projection of a message onto the hidden subspace $U_L$ is $\sqrt{N/n}$, and it must be below the margin $\tau\approx N^{-5/4}/400$, which gives $n\gtrsim N\cdot N^{5/2}=N^{7/2}$ (indeed we get $n=1.1\cdot10^6N^{7/2}$ to accommodate small values of $N$).
We have not tried to optimize this exponent.

\section{OFW with \texorpdfstring{$m$}{m} LMO calls per round}\label{sec:ub}

We design a generalization of OFW \citep{hazan2012projection} with $m$ $\LMO$ calls per round (\Cref{alg:mofw-opt}). We show that its regret is $O(LD\,T^{3/4}m^{-1/4})$ when $m=O(T)$ (\cref{cor:pot}), and sketch the potential-function proof, which follows \OPT. We then give an anytime tuning (which does not depend on~$T$) with the same rate (\cref{cor:pot-any}). 
The minimax rate $O(LD\max\{T^{3/4}m^{-1/4},\sqrt T\})$ is attained for all~$m$ and~$T$ by a tuning capped at $m=T$ (\cref{cor:pot-cap}). 

In this section rounds are numbered $t=1,\dots,T$, the starting point is $x_1\in K$, $G_t:=\sum_{s=1}^tg_s$, and for a fixed comparator $x_\star\in K$ we write $\Reg_T:=\sum_{t=1}^T\ip{g_t}{x_t-x_\star}$. 
\begin{assumption}\label{ass:ub}
The body $K$ has diameter at most $D$, the loss vectors satisfy $\norm{g_t}\le L$ for all $t$, and the LMO may break ties arbitrarily.
\end{assumption}
These are the conventions of \OPT, so that for $m=1$ every statement reduces to the corresponding statement there. 

\begin{algorithm}[!h]
\caption{$m$-step Online Frank--Wolfe}\label{alg:mofw-opt}
\begin{algorithmic}[1]
\Require $x_1\in K$, $\eta>0$, $\sigma\in(0,\tfrac12]$
\For{$t=1,\dots,T$}
  \State play $x_t$, observe $g_t$; set $z_{t,0}\leftarrow x_t$
  \For{$k=1,\dots,m$}
    \State $\dir_{t,k}\leftarrow\eta\,G_t+(z_{t,k-1}-x_1)$
    \State $v_{t,k}\leftarrow\LMO(\dir_{t,k})$
    \State $z_{t,k}\leftarrow(1-\sigma)\,z_{t,k-1}+\sigma\,v_{t,k}$
  \EndFor
  \State $x_{t+1}\leftarrow z_{t,m}$
\EndFor
\end{algorithmic}
\end{algorithm}

In \Cref{alg:mofw-opt}, the query $\dir_{t,k}$ is the gradient at $z_{t,k-1}$ of $F_t(x)=\eta\ip{G_t}x+\frac12\norm{x-x_1}^2$, so a round consists of $m$ Frank--Wolfe steps with constant step size $\sigma$ on $F_t$, started at $x_t$. Every iterate is a convex combination of $x_1$ and of returned atoms, and every query is a linear combination of $x_1$ and of past gradients and atoms, so the method is a span method and \cref{thm:span} applies to it. 
\cref{alg:mofw-opt} enjoys a regret upper bound, stated in \Cref{cor:pot}, that matches the lower bound of \cref{thm:span} for $m=O(T)$.

\begin{theorem}[Upper bound for OFW with $m$ calls per round]\label{cor:pot}
Under \Cref{ass:ub},
for every $m,T\ge 1$ with $mT\ge12$, and for every $x_\star\in K$,
$m$-step Online Frank--Wolfe (\cref{alg:mofw-opt}) with parameters
\begin{equation}\label{eq:def-param-OFW}
    \eta=\frac{D}{2L}\Bigl(\frac3T\Bigr)^{3/4}m^{1/4},\qquad\sigma=\sqrt{\frac{3}{mT}} \, , 
\end{equation}
has regret
\[
\Reg_T\ \le\ \frac{4}{3^{3/4}}\,LD\,T^{3/4}m^{-1/4}+\frac{4}{3^{1/4}}\,LD\,(mT)^{1/4}.
\]
For $m=1$ the bound $\Reg_T\le4\cdot3^{-3/4}LD\,T^{3/4}$ holds without the second term.
\end{theorem}

The proof is in Appendix~\ref{sec:proof:ofw-fixed}.
Note that, by the standard linearization trick for OCO, the theorem extends to convex $L$-Lipschitz losses $f_t$: run the algorithm on the subgradients $g_t \in \partial f_t(x_t)$, and use that, by convexity, $f_t(x_t)-f_t(u)\le\ip{g_t}{x_t-u}$ for all $u\in K$.

\paragraph{The potential-function proof.} 
Let $y_{t+1}:=\argmin_{y\in K}F_t(y)$, with $y_1:=x_1$, be the exact follow-the-regularized-leader (FTRL) iterates. 
\OPT prove the constant $4\cdot3^{-3/4}$ for $m=1$ by tracking the potential
\begin{multline*}
 \phi_t:=\sum_{s=1}^{t}\ip{g_s}{x_s-y_{t+1}}-\frac{1}{2\eta}\norm{y_{t+1}-x_1}^2\\
 +\frac{1}{6\eta}\norm{x_{t+1}-y_{t+1}}^2.
\end{multline*}
Their proof is based on upper bounding the potential increase $\phi_t - \phi_{t-1}$ for every round~$t\ge 1$ (with $\phi_0=0$).
The LMO step enters their proof only through the squared distance $\norm{x_{t+1}-y_{t+1}}^2$, %
and is handled with the following one-step contraction argument.
For a step $z'=(1-\sigma)z+\sigma v$, where $v$ minimizes the linearization of $F_t$ at $z$, 
the optimality of $v$ and of $y_{t+1}$ 
yields the one-step contraction
\begin{equation}\label{eq:one-step}
 \norm{z'-y_{t+1}}^2\le(1-2\sigma)\norm{z-y_{t+1}}^2+\sigma^2\norm{z-v}^2 .
\end{equation}
For $m=1$ it is applied once, to $z=x_t$ and $z'=x_{t+1}$; the rest of the proof involves only $x_t$ and the exact iterates $y_t,y_{t+1}$ and does not depend on how $x_{t+1}$ was produced. With $m$ calls, the $m$ steps of round $t$ are LMO steps on the same $F_t$, towards the same $y_{t+1}$, so each of them satisfies \eqref{eq:one-step} with $(z,v,z')=(z_{t,k-1},v_{t,k},z_{t,k})$. Chaining the $m$ inequalities, which requires $1-2\sigma\ge0$ and is the reason for $\sigma\le\frac12$, gives
\begin{multline}\label{eq:m-steps}
 \norm{x_{t+1}-y_{t+1}}^2\le(1-2\sigma)^m\norm{x_t-y_{t+1}}^2\\
 +\sigma^2\sum_{k=1}^m(1-2\sigma)^{m-k}\norm{z_{t,k-1}-v_{t,k}}^2 .
\end{multline}
A round with $m$ LMO calls thus contracts the squared distance by the factor $1-u$, where
\begin{equation}\label{eq:u}
 u:=1-(1-2\sigma)^m\in(0,1],
\end{equation}
exactly as one LMO call contracts it by $1-2\sigma$. 
The error of step $k$ is damped by the $m-k$ contractions that follow it. Replacing \eqref{eq:one-step} by \eqref{eq:m-steps} in the proof of \OPT, and $2\sigma$ by $u$ in their constants, gives the per-round inequality (\cref{lem:pot})
\begin{multline*}
 \phi_t-\phi_{t-1}\ \le\ \frac{8\eta}{3u}\norm{g_t}^2\\
 +\frac{\sigma^2}{6\eta}\sum_{k=1}^m(1-2\sigma)^{m-k}\norm{z_{t,k-1}-v_{t,k}}^2 .
\end{multline*}
Summing over $t$ and using the optimality of $y_{T+1}$,
$\norm{z_{t,k-1}-v_{t,k}}\le D$ and $\sum_{k=1}^m(1-2\sigma)^{m-k}=u/(2\sigma)$ gives, for every $x_\star\in K$,
\[
 \Reg_T\ \le\ \frac{8\eta L^2T}{3u}+\frac{\sigma u\,D^2T}{12\eta}+\frac{D^2}{2\eta}
\]
(\cref{thm:pot}), the same bound as in \OPT for single-call OFW, with $2\sigma$ replaced by~$u$. 
Since $u\approx2m\sigma$ at first order, a round of $m$ LMO steps tracks $y_{t+1}$ like a single step of size $m\sigma$ (same first error term), while its atom error for the second term
pays a factor~$m$ instead of~$m^2$.
For $u=2m\sigma$ the three terms would be $2$, $1$ and $1$ times $3^{-3/4}LD\,T^{3/4}m^{-1/4}$, whence the constant $4\cdot3^{-3/4}$ for every $m$.
The lower-order term $4\cdot3^{-1/4}LD(mT)^{1/4}$ accounts for $u\ne2m\sigma$ through $1/u\le1/(2m\sigma)+1$. 

\paragraph{Anytime tuning.} An adaptation of this proof (\Cref{sub:pot-anytime}) 
handles time-varying parameters, 
that is, \Cref{alg:mofw-opt} with $\eta,\sigma$ replaced in round $t-1$ by
\begin{equation}\label{eq:def-param-anytime}
\eta_t=\frac{D}{2L}\Bigl(\frac3t\Bigr)^{3/4} m^{1/4},\quad
\sigma_t=\min\Bigl\{\frac12,\sqrt{\frac{3}{mt}}\Bigr\} 
\end{equation}
to compute $x_t$.
This method does not depend on~$T$.

\begin{theorem}[Anytime $m$-step OFW, Upper bound]\label{cor:pot-any}
Under \Cref{ass:ub},
for every $m, T\ge1$ and $x_\star\in K$,
\cref{alg:mofw-opt} with the time-varying parameters~\eqref{eq:def-param-anytime} has regret
\[
 \Reg_T\ \le\ \frac{5}{3^{3/4}}\,LD\,T^{3/4}m^{-1/4}+57\,LD\,(mT)^{1/4}.
\]
\end{theorem}

The lower-order constant is crude. %
For $m=1$ and the schedule $\sigma_t=\min\{1,\sqrt{3/t}\}$, \OPT prove $\Reg_T\le5\cdot3^{-3/4}LD\,T^{3/4}$ for every $T$, without the lower-order term.

\paragraph{The regime $m\ge T$.} The above tunings of Theorems~\ref{cor:pot} and~\ref{cor:pot-any} are suboptimal for $m\ge T$ since the term $(mT)^{1/4}$ dominates.
In this case, replacing $m$ with $\min\{m, t\}$ in parameter $\eta_t$, 
(and with $\min\{m, T\}$ in the fixed-horizon parameter $\eta=\eta_T$),
that is setting
\begin{equation}\label{eq:def-param-OFW-capped}
    \tilde \eta_t = \frac{D}{2L}\Bigl(\frac3t\Bigr)^{3/4}\min\{m,t\}^{1/4} ,
\end{equation}
recovers the minimax $\sqrt{T}$ rate of OCO with projection. 
With $m\ge T$ LMO calls per round, the OFW iterate tracks the exact FTRL iterate closely enough that the method inherits its $\sqrt T$ regret.

\begin{corollary}[$m$-step OFW with capped tuning]\label{cor:pot-cap}
\textbf{(i) Fixed-horizon case:}
Under \Cref{ass:ub},
for every $m, T\ge1$ with $mT\ge 12$, and every $x_\star\in K$,
\cref{alg:mofw-opt} 
with parameters $\sigma$ as in~\eqref{eq:def-param-OFW}
and $\eta = \tilde \eta_T$ as in~\eqref{eq:def-param-OFW-capped}
has regret
\[
\Reg_T
\ \le\ 4.8\,LD\max\{T^{3/4}m^{-1/4},\sqrt T\}.
\]
\textbf{(ii) Anytime case:}
Under \Cref{ass:ub}, for every $m, T\ge1$, and every $x_\star\in K$,
\cref{alg:mofw-opt} with anytime parameters $\sigma_t$ as in~\eqref{eq:def-param-anytime}
and $\eta_t = \tilde \eta_t$ as in~\eqref{eq:def-param-OFW-capped}
has regret
\[
\Reg_T
\ \le\ 60 \,LD\max\{T^{3/4}m^{-1/4},\sqrt T\}.
\]
\end{corollary}

The complete proof is in \Cref{sec:proof-capped-tuning}.
Note that the constant~$60$ of the anytime bound is crude.

\section{Conclusion}\label{sec:conclusion}

In online convex optimization, the $T^{3/4}$ regret of OFW is the price of accessing a high-dimensional body through $O(1)$ linear optimization calls per round: one known simplex in dimension $2m(T-1)+2$ enforces it on every span method; and a random rotation of the same simplex in dimension $\gtrsim(mT)^{7/2}$, with a small ball as decoy, enforces it on every learner. With $m$ calls per round the minimax regret is $\Theta(LD\max\{T^{3/4}m^{-1/4},\sqrt T\})$, attained by a generalization of OFW with $m$ LMO calls per round. 

\label{end:main}

\subsection*{Acknowledgments}
This work took place in the context of the associate team 4TUNE within \href{https://project.inria.fr/inriacwi/4tune/}{CWI-Inria international lab}. 
Julien Weibel and Adrien Taylor are supported by the European Union (ERC grant CASPER 101162889). 
Views and opinions expressed are however those of the author(s) only and do not necessarily reflect those of the European Union or the European Research Council. Neither the European Union nor the granting authority can be held responsible for them.
The French government also partly funded this work under the management of Agence Nationale de la Recherche as part of the ``France 2030'' program, reference ANR-23-IACL-0008 (PR[AI]RIE-PSAI)
and reference ANR-23-IACL-0006 (MIAI Cluster).

\subsection*{AI use statement}

Generative AI tools were used in this work to draft and revise the text, to carry out numerical and symbolic verification of the constants, and to search the literature. All AI-assisted proofs were checked line by line by the authors, and the literature search was complemented by a manual survey. We take responsibility for the final content of this work, including text, claims, and artifacts produced with the aid of generative AI.

\clearpage
\appendix
\thispagestyle{empty}
\onecolumn
\aistatstitle{An $\Omega(T^{3/4})$ lower bound for projection-free online learning with linear optimization oracles:\\ Supplementary Materials}

\section{Proofs for Section~\ref{sec:span}: span methods}
\label{app:span}

This section contains the complete proofs of all the results in \Cref{sec:span}, and in particular \Cref{thm:span}.
Throughout this section, the instance we consider is the one defined in \cref{sec:span-instance}: unshrunk atoms $\bar w_j$ in $\R^{2N}$, the simplex~$\bar K$, the canonical losses~$\bar g^{(\kappa)}$ with the pattern~\eqref{eq:canon}, the schedule $\kappa_t=mt$, and the oracle returning the first minimizer in the list $0,\bar w_0,\dots,\bar w_{N-1}$.

\subsection{Preliminary bounds on the profile \texorpdfstring{$e_j$}{e}}\label{app:profile}

Recall from \cref{sec:span-instance} that $h_k=\sqrt{2k+1}$, $\hat h_k=h_k/(2\sqrt N)$, and that the profile is $e_j=\hat h_j$ in the lower half $j<N/2$ and $e_j=1-\hat h_{N-1-j}$ in the upper half $j\ge N/2$, with increments $\delta_j=e_j-e_{j-1}$ and noise sizes $f_j=\min(e_j,1-e_j)$.

To prove preliminary bounds on the profile $e_j$, we will need the following auxiliary sequences
\[
a_j=\min(j,N-j),\qquad b_j=\min(j,N-1-j),
\]
\[
\psi_0=2,\qquad \psi_k=h_k+h_{k-1}\ \ (k\ge1),\qquad \Psi_j=\psi_{a_j}.
\]
Since $h_k^2-h_{k-1}^2=2$, we have $\psi_k=2/(h_k-h_{k-1})$ for $k\ge1$.

\begin{lemma}[Properties of the profile $e_j$]\label{lem:profile}
The following statements hold:
\begin{enumerate}[label=(\alph*),leftmargin=*,topsep=0pt,itemsep=1pt]
\item Monotonicity: The profile is strictly increasing and lies in $(0,1)$; in particular all $\delta_j$ and $f_j$ are positive.
\item Noise sizes: $\sqrt N\,f_j=h_{b_j}/2$.
\item Increments: $\sqrt N\,\delta_j\ge1/\Psi_j$.
\item Index functions: for $0\le i\le N-2$, $a_{i+1}\ge1$ and $b_i\in\{a_{i+1}-1,\,a_{i+1}\}$.
\item Gaps: $h_0=1$ and $0<h_k-h_{k-1}\le1$ for $k\ge1$.
\item Distances: $e_j+f_j\le1$ for every $j$, and $e_j-e_i+f_i+f_j\le1$ for $i<j$.
\end{enumerate}
\end{lemma}
 
Facts (b) and (c) express the noise sizes and increments, after multiplication by $\sqrt N$, through the explicit numbers~$h_k$. Together with (d) and (e), they are the inputs of the two recursions that lower bound the per-round costs: \cref{lem:qrec} below, used for both instances through \cref{lem:cbar,lem:level}, and \cref{lem:cancel} for the hidden instance. Fact~(a) gives the simplex structure and fact~(f) the diameter bound of span-method instance. The proof of the lemma is an elementary case analysis.

\begin{proof}[Proof of \cref{lem:profile}]

The facts are checked one at a time. 

\step{(a) Monotonicity.} In the lower half $j<N/2$, we have $2j+1\le N$, so $e_j=\hat h_j\le\frac12$. In the upper half $j\ge N/2$, we have $2(N-1-j)+1\le N-1$, so $\hat h_{N-1-j}<\frac12$ and $e_j=1-\hat h_{N-1-j}>\frac12$. The profile is increasing within each half, since $\hat h_k$ is increasing in $k$, and it is increasing across the middle because every lower value is at most~$\frac12$ and every upper value is above~$\frac12$. Finally $e_0=\hat h_0>0$ and $e_{N-1}=1-\hat h_0<1$. Hence, all $\delta_j$ and $f_j$ are positive, since the profile is strictly increasing with $e_{-1}=0$ and lies in $(0,1)$.

\step{(b) Noise sizes.} By the bounds of (a), the noise size is the hatted term in both halves:
\[
f_j=\min(e_j,1-e_j)=\begin{cases}\hat h_j,& j<N/2,\\ \hat h_{N-1-j},& j\ge N/2,\end{cases}
\]
which is $\hat h_{b_j}$, because $b_j=j$ in the lower half and $b_j=N-1-j$ in the upper half. Multiplying by $\sqrt N$ gives~(b).

\step{(c) Increments.} We show $1/\delta_j\le\sqrt N\,\Psi_j$ by distinguishing four cases according to the value of $j$.
\begin{itemize}[leftmargin=*,itemsep=1pt]
\item $j=0$: We have $\delta_0=\hat h_0=1/(2\sqrt N)$, so $1/\delta_0=2\sqrt N=\sqrt N\,\psi_0$ as $a_0=0$.
\item $1\le j<N/2$: Both $j$ and $j-1$ are in the lower half, $\delta_j=(h_j-h_{j-1})/(2\sqrt N)$, so $1/\delta_j=\sqrt N\,\psi_j$ as $a_j=j$.
\item $j>N/2$: Then we have $\delta_j=\hat h_{N-j}-\hat h_{N-1-j}$. This is clear when $j-1$ is also in the upper half. Otherwise $N$ is odd and $j=(N+1)/2$, and it still holds because $e_{j-1}=\hat h_{(N-1)/2}=\frac12=1-\hat h_{N-j}$, since $j-1 = (N-1)/2 = N-j$. So in both cases, we have $1/\delta_j=\sqrt N\,\psi_{N-j}$ as $a_j=N-j$.
\item $j=N/2$ ($N$ even): Here $\delta_j=1-2\hat h_{j-1}=1-\sqrt{(N-1)/N}$, and as $a_j=N/2$, we have
\[
\frac1{\delta_j}=\sqrt N\big(\sqrt N+\sqrt{N-1}\big)\ \le\ \sqrt N\big(\sqrt{N+1}+\sqrt{N-1}\big)=\sqrt N\,\psi_{N/2}.
\]
\end{itemize}

\step{(d) Index functions.} For $0\le i\le N-2$, $a_{i+1}=\min(i+1,N-1-i)\ge1$. If $i+1\le N-1-i$, then $a_{i+1}=i+1$ and $b_i=i$. Otherwise $N-1-i\le i$, and $a_{i+1}=N-1-i=b_i$.

\step{(e) Gaps of $h$.} We have $h_0=1$. For $k\ge1$ the gap is $h_k-h_{k-1}=2/(h_k+h_{k-1})$, which is at most $2/(1+\sqrt3)<1$.

\step{(f) Distances.} For every $j$, we have $e_j+f_j=\min(2e_j,1)\le1$. For $i<j$ there are three cases. If both~$e_i$ and~$e_j$ are at most $\frac12$, the sum $e_j-e_i+f_i+f_j$ equals $2e_j\le1$. If both are at least $\frac12$, it equals $2-2e_i\le1$ (recall $e_i < 1$). If $e_i\le\frac12\le e_j$, it equals $1$. In all three cases, the sum is at most $1$.
\end{proof}

\subsection{Proof of \cref{lem:ties}}
\label{app:proof_ties}

\begin{proof}[Proof of \cref{lem:ties}]
By linearity, it suffices to check $\ip{d}{\bar w_j}=\ip{d}{\bar w_J}$ for $j\ge J$ when $d$ is one of the generating vectors. For $d=0$ both sides vanish. For $d=\bar w_i$ with $i<J\le j$, the martingale identity gives $\ip{\bar w_i}{\bar w_j}=e_i=\ip{\bar w_i}{\bar w_J}$. For $d=g_s=\bar g^{(\kappa_s)}$ with $s\le t-1$, we have $\kappa_s\le\kappa_{t-1}\le J\le j$, so \eqref{eq:canon} gives $\ip{g_s}{\bar w_j}=-\bar c_{\kappa_s}=\ip{g_s}{\bar w_J}$. Hence all atoms of index at least $J$ have the same value $\ip{d}{\bar w_J}$, which proves the first statement of the lemma.

In particular, for any query $d\in\spn(0,\bar w_0, \dots, \bar w_{J-1},g_0,\dots, g_{t-1})$,
the first minimizer of $\ip d\cdot$ over the list $0,\bar w_0,\dots,\bar w_{N-1}$ is either $0$ or an atom $\bar w_i$ with index $i\le J$.
In both cases $\LMO(d)\in\{0,\bar w_0,\dots,\bar w_J\}$.
\end{proof}

\subsection{Proof of \cref{prop:span}}
\label{app:proof_propspan}

\begin{proof}[Proof of \cref{prop:span}]
We first show by induction over the queries that the $k$-th query of round $t\ge1$ returns~$0$ or an atom of index at most $J:=m(t-1)+k-1$. Indeed, by the induction hypothesis the atoms returned before this query have index below $J$, so $d_{t,k}\in\spn(0,\bar w_0,\dots,\bar w_{J-1},g_0,\dots,g_{t-1})$ since the learner is assumed to be a span method,
and applying \cref{lem:ties} (note that $J\ge\kappa_{t-1} = m(t-1)$) completes the induction step.

In particular, every atom returned before the play of round $t$ has index at most $m(t-1)+m-1=\kappa_t-1$. Consequently, $x_t$ is a convex combination of $0$ and of atoms $\bar w_j$ of index $j<\kappa_t$, which all have zero inner product with the loss vector $g_t=\bar g^{(\kappa_t)}$ by \eqref{eq:canon}. Hence we have $\ip{g_t}{x_t}=0$ for all $t<T$.

Finally, the cumulative loss $u\mapsto\sum_{t<T}\ip{g_t}{u}$ is linear, so its minimum over $\bar K$ is attained at a vertex: it is $0$ at the origin and $-\sum_{t:\,\kappa_t\le j}\bar c_{\kappa_t}$ at atom $\bar w_j$ by \eqref{eq:canon}. As every $\bar c_\kappa$ is positive, the minimum is attained at $j=N-1$ and, since $\kappa_t\le N-1$ for all $t<T$, equals $-\sum_{t<T}\bar c_{\kappa_t}$. Therefore $\Reg_T=\sum_{t<T}\bar c_{\kappa_t}$ for all span methods.
\end{proof}

\subsection{Proof of \cref{lem:cbar}}
\label{app:proof:cbar}

We first compute the distance from the walk position $M_\kappa$ to the span of the head atoms, by eliminating the unknowns of a least-squares problem one at a time along the chain. Each elimination uses the same identity: for $x,y>0$ define the \emph{parallel sum}
\[
x\psum y=\frac{xy}{x+y},\qquad\text{equivalently}\qquad \frac1{x\psum y}=\frac1x+\frac1y ,
\]
the formula for two electrical resistances in parallel. It is increasing in $x$ and in $y$, and it is the value of a one-variable quadratic problem:
\begin{equation}\label{eq:par}
\min_{t\in\R}\big(x\,t^2+y\,(t-1)^2\big)=x\psum y .
\end{equation}

\begin{lemma}[Distance of $M_\kappa$ to the head span]\label{lem:qrec}
For $0\le\kappa\le N-1$, we have
\[
\dist\big(M_\kappa,\spn(\bar w_0,\dots,\bar w_{\kappa-1})\big)^2
=q_\kappa:=\min_{Z_0,\dots,Z_{\kappa-1}}\ \Big[\ \sum_{i=0}^{\kappa}\delta_i\,Z_i^2+\sum_{s=0}^{\kappa-1}f_s\,(Z_s-Z_{s+1})^2\ \Big]\qquad\text{with }Z_\kappa=1,
\]
and $q_\kappa\ge1/(2\sqrt N)$. In particular, we have
\[ \dist(M_\kappa,\spn(\bar w_0,\dots,\bar w_{\kappa-1}))\ge(4N)^{-1/4} . \]
\end{lemma}

\begin{proof}
\firststep{Step 1: The equality.} By definition, the squared distance $\dist(M_\kappa,\spn(\bar w_0,\dots,\bar w_{\kappa-1}))^2$ is equal to $\min_{y_0,\dots,y_{\kappa-1}}\norm{M_\kappa-\sum_{s<\kappa}y_s\bar w_s}^2$. We write the vector inside the norm in the (orthonormal) basis $(u_i,\nu_i)_{i<N}$ and replace the coefficients $y_s$ by the cumulative quantities $Z_i=1-\sum_{s=i}^{\kappa-1}y_s$ for $0\le i\le\kappa$, so that $Z_\kappa=1$ and $y_s=Z_{s+1}-Z_s$. This change of variables is a bijection. The coordinate along $u_i$ collects $\sqrt{\delta_i}$ from $M_\kappa$ and $-y_s\sqrt{\delta_i}$ from each atom $\bar w_s$ with $s\ge i$, so it equals $\sqrt{\delta_i}\,Z_i$. The coordinate along $\nu_s$ is $-y_s\sqrt{f_s}=\sqrt{f_s}\,(Z_s-Z_{s+1})$. Summing the squares of those coordinates gives the equality of the lemma.

\step{Step 2: The sequence $(q_k)_{k<N}$ satisfies a recursion formula.} 
The problem defining~$q_{k+1}$ consists of the problem for~$q_k$, with $Z_k$ now free, plus the two terms $\delta_{k+1}Z_{k+1}^2=\delta_{k+1}$ and $f_k(Z_k-1)^2$. 
If $Z_k$ is fixed to a value $t$ instead of $1$ in the definition of $q_k$, the minimum becomes $q_k\,t^2$ by homogeneity. 
Thus, minimizing first over $Z_0,\dots,Z_{k-1}$ and then over $Z_k$ in the definition of $q_{k+1}$ gives
\begin{align*}
q_{k+1}
& = \delta_{k+1} + \min_{Z_k} 
    \Big[\ f_k (Z_k - 1)^2 + \min_{Z_0,\dots,Z_{k-1}}  \Big[\     
            \sum_{i=0}^{k}\delta_i\,Z_i^2+\sum_{s=0}^{k-1}f_s\,(Z_s-Z_{s+1})^2\ \Big] \ \Big] \\
& = \delta_{k+1} + \min_{Z_k} 
    \Big[\ f_k (Z_k - 1)^2 + q_k Z_k^2\ \Big] ,
\end{align*}
and using~\eqref{eq:par} gives the recursion
\[
q_0=\delta_0,\qquad q_{k+1}=\delta_{k+1}+\big(q_k\psum f_k\big).
\]

\step{Step 3: The recursion stays above $1/(2\sqrt N)$.} Let $Q_k=\sqrt N\,q_k$. We show by induction that $Q_k\ge\frac12$ for all $k<N$. For $k=0$, we have $Q_0=\sqrt N\,\delta_0=\frac12$. 
Let $k>0$, and assume that $Q_{k-1} \ge \frac12$. Multiplying the recursion by $\sqrt N$ and using \cref{lem:profile}(b,c), we get
\[
Q_{k}\ \ge\ \frac1{\Psi_{k}}+\Big(Q_{k-1}\psum\frac{h_b}{2}\Big),\qquad b=b_{k-1}.
\]
Since the parallel sum $\cdot\psum\cdot$ is increasing in its two arguments, and as $Q_{k-1}\ge\frac12$, we get
\[
Q_k\ \ge\ \frac1{\Psi_k}+\Big(\frac12\psum\frac{h_b}{2}\Big)=\frac1{\Psi_k}+\frac{h_b}{2(h_b+1)} .
\]
The right side is at least $\frac12$ exactly when $\Psi_k\le2(h_b+1)$, and this is where the choice of profile $(e_j)_{j<N}$ enters (through \cref{lem:profile}). We have $\Psi_k=h_a+h_{a-1}$ with $a=a_k\ge1$, and \cref{lem:profile}(d) gives $b\ge a-1$, so $h_b\ge h_{a-1}$. \Cref{lem:profile}(e) then gives
\[
\Psi_k=h_a+h_{a-1}\ \le\ 2h_{a-1}+1\ \le\ 2h_b+2 .
\]
Hence, we have $Q_k\ge\frac12$ for every $k<N$, that is, $q_k \ge1/(2\sqrt N)$.
This concludes the proof of the lemma.
\end{proof}

Using \Cref{lem:cbar}, we can now prove \cref{lem:cbar}.

\begin{proof}[Proof of \cref{lem:cbar}]
Let $\kappa$ with $0\le\kappa\le N-1$ be fixed. Write $\bar g=\bar g^{(\kappa)}$ and $\bar c=\bar c_\kappa$.

\firststep{The equality.} Let $y=\sum_{j\ge\kappa}\lambda_j\bar w_j$ with $\sum_{j\ge\kappa}\lambda_j=1$ be a point of the affine hull $\aff(\bar w_{\kappa},\dots,\bar w_{N-1})$ and let $z\in\spn(\bar w_0,\dots,\bar w_{\kappa-1})$. By~\eqref{eq:canon}, we have $\ip{\bar g}{y-z}=-\bar c\sum_{j\ge\kappa}\lambda_j = -\bar c$, and since $\norm{\bar g}=1$ the Cauchy--Schwarz inequality gives $\norm{y-z}\ge\bar c$. Conversely, since $\bar g^{(\kappa)}=-\bar c_\kappa\,\bar WA^{-1}\indVec_\kappa$, writing $\bar g=\sum_j\beta_j\bar w_j$ with $\beta=-\bar c\,A^{-1}\indVec_\kappa$, and using~\eqref{eq:canon} we get
\[
1=\ip{\bar g}{\bar g}=\sum_j\beta_j\ip{\bar g}{\bar w_j}=-\bar c\sum_{j\ge\kappa}\beta_j .
\]
So $\bar c=\norm{\bar c\,\bar g}=\norm{y^*-z^*}$ where $y^*=-\bar c\sum_{j\ge\kappa}\beta_j\bar w_j\in \aff(\bar w_{\kappa},\dots,\bar w_{N-1})$ since its weights $(-\bar c\beta_j)_{j\ge\kappa}$ sum to $1$, and $z^*=\bar c\sum_{j<\kappa}\beta_j\bar w_j \in\spn(\bar w_0,\dots,\bar w_{\kappa-1})$. Hence we get the equality in the lemma.

\step{The first inequality.} Every tail atom $\bar w_j$ with $j\ge\kappa$ is $\bar w_j=M_\kappa+a_j$ where $a_j=\sqrt{f_j}\,\nu_j + \sum_{\kappa<i\le j}\sqrt{\delta_i}\,u_i$. 
So any $y=\sum_{j\ge\kappa}\lambda_j\bar w_j$ with $\sum_{j\ge\kappa}\lambda_j=1$ in the affine hull $\aff(\bar w_{\kappa},\dots,\bar w_{N-1})$ can be written as $M_\kappa+a$ with $a=\sum_{j\ge\kappa}\lambda_ja_j$. 
The vector $a$ is orthogonal to $E_\kappa=\spn(u_0,\dots,u_\kappa,\nu_0,\dots,\nu_{\kappa-1})$, which contains $M_\kappa$ and the head atoms. Thus for $z\in\spn(\bar w_0,\dots,\bar w_{\kappa-1})$, we have $\norm{y-z}^2=\norm{M_\kappa-z}^2+\norm a^2\ge\norm{M_\kappa-z}^2$. Taking the infimum over $y$ and $z$ gives $\bar c_\kappa\ge\dist(M_\kappa,\spn(\bar w_0,\dots,\bar w_{\kappa-1}))$.

\step{The second inequality.} This is \cref{lem:qrec}.
\end{proof}

\subsection{Proof of \cref{thm:span}}
\label{app:proof_thmspan}

\begin{proof}[Proof of \cref{thm:span}]
Consider the instance of \cref{sec:span-instance}. It has the stated form: $\norm{g_t}=\norm{\bar g^{(\kappa_t)}}=1$ for all $t$, $\bar K\subset\R^{2N}$, and $\diam\bar K\le1$ by \cref{lem:profile}(f), since by the martingale identity two atoms are at squared distance $\norm{\bar w_j-\bar w_i}^2=e_j-e_i+f_i+f_j\le1$ for $i<j$, and an atom $\bar w_j$ is at distance $\sqrt{e_j+f_j}\le1$ from the origin $x_0$.

By \cref{prop:span}, every span method making at most $m$ oracle calls per round has regret $\Reg_T=\sum_{t<T}\bar c_{\kappa_t}$, and by \cref{lem:cbar} each $\bar c_{\kappa_t}$ is at least $(4N)^{-1/4}$. Hence we get $\Reg_T\ge T(4N)^{-1/4}\ge T^{3/4}(4m)^{-1/4}$, where the last inequality comes from $N=m(T-1)+1\le mT$. This proves the theorem.

Finally, for $L,D>0$, the scaled instance $(D\bar K,(Lg_t)_{t<T})$ has diameter at most $D$ and loss vectors of norm $L$. A span method run on it is, after the changes of variables $x\mapsto x/D$ and $d\mapsto d/L$, a span method run on the original instance with the same oracle answers up to the factor $D$, so its regret is $LD$ times the original one. This proves the statement after \cref{thm:span} for scaled instances.
\end{proof}

\section{Proof of Theorem~\ref{thm:all}: every learner}\label{app:all}

This section contains the complete proofs of all the results in \Cref{sec:all}, and in particular \Cref{thm:all} and \Cref{cor:all}.
We start by proving \cref{cor:all}, which is a direct corollary of \Cref{thm:all} by choosing a learner-dependent rotation of the instance's random frame.

\begin{proof}[Proof of \cref{cor:all}]
Fix a learner making at most $m$ oracle calls per round. In \cref{thm:all} the expectation is over the random frame and over the learner's randomness~$\xi$, so the average over the frame of $\E_\xi[\Reg_T]$ is at least $0.23\,T\,N^{-1/4}$ (recall the instance's randomness is independent of the learner's randomness~$\xi$).
Hence, some realization of the frame, that is, some fixed instance depending on the learner (but not on~$\xi$), satisfies $\E_\xi[\Reg_T]\ge0.23\,T\,N^{-1/4}$. Scaling this instance by $D/(\sqrt5/2)$ in space and by $L$ in the losses gives $\diam K\le D$ and $\norm{g_t}=L$, and multiplies the regret by $2LD/\sqrt5$. Since $0.23\cdot2/\sqrt5>0.2$, the claim follows.
\end{proof}

The rest of this section is devoted to the proof of \cref{thm:all}. As discussed directly after its statement, it suffices to treat deterministic learners: the instance distribution does not depend on the learner nor on its internal randomness~$\xi$, so conditionally on~$\xi$ a randomized learner behaves as a deterministic one, and both bounds of the theorem, once proved for every fixed value of $\xi$, therefore hold unconditionally (Yao's principle). \emph{From now on the learner is deterministic.}

Throughout this section, the instance we consider is the hidden instance of \cref{sec:all-instance}: the random frame $(u_0,\nu_0,\dots,u_{N-1},\nu_{N-1})$ in $\R^n$, the shrunk atoms $w_j=\rho_jM_j+\sqrt{f_j}\,\nu_j$ with $\rho_j=1-j/(2(N-1))$ and $\gamma=\rho_j-\rho_{j+1}=1/(2(N-1))$, the body $K=\conv(\{0\}\cup\{w_j\}\cup B_r)$ with $r=\frac1{40}N^{-1/4}$, the oracle returning the first minimizer in the list $0,w_0,\dots,w_{N-1},-r\,d/\norm d$, the vectors $g^{(\kappa)}=-(M_\kappa-\Pi_\kappa M_\kappa)/c_\kappa$ and the losses $g_t=g^{(\kappa_t)}$, with $\kappa_t=mt$. The proof has five steps, one per subsection: build the losses and the levels (\cref{sec:levels}); prove four deterministic facts about the geometry of $K$ (\cref{sec:lemmas}); introduce a clean oracle that ignores the hidden directions, and show that it agrees with the true oracle as long as the learner's messages are nearly orthogonal to them (\cref{sec:clean}); show that the messages are indeed nearly orthogonal to the hidden directions, with high probability (\cref{sec:hidden}); and put the pieces together (\cref{sec:proof}).

\subsection{The losses and the process levels}\label{sec:levels}

We first prove the loss vectors $g^{(\kappa)}$ have a similar pattern as those of the span-method instance from \cref{sec:span-instance}.

\begin{lemma}[Structure of the losses]\label{lem:F1}
Let $0\le \kappa \le N-1$, and let $E_\kappa=\spn(u_0,\dots,u_\kappa,\nu_0,\dots,\nu_{\kappa-1})$. The vector $g^{(\kappa)}$ is a unit vector of $E_\kappa$, and we have
\[
\ip{g^{(\kappa)}}{w_j}=\begin{cases}0,& j<\kappa,\\[2pt] -\rho_j\,c_\kappa,& j\ge\kappa.\end{cases}
\]
In particular, we have $\ip{g^{(\kappa)}}{b}\ge-r$ for every $b\in B_r$.
\end{lemma}

This means the loss $g^{(\kappa)}$ cannot be exploited with atoms before $\kappa$, it rewards all atoms from $\kappa$ on by $c_\kappa$ times their shrinkage factor $\rho_j$, hence by an amount between $c_\kappa /2$ and $c_\kappa$, and it rewards the comparator by exactly $c_\kappa/2$, since $\rho_{N-1}=\frac12$.
As discussed in \cref{sec:all-instance}, the loss vector $g^{(\kappa)}$ is in $E_\kappa$ whereas the loss vectors were in the span of all the atoms in the span-method instance of \cref{sec:span-instance}, which would reveal the frame.

\begin{proof}
\firststep{Step 1: The earlier atoms.} The walk position $M_\kappa$ and the atoms $w_s$ with $s<\kappa$ are built from the vectors $u_0,\dots,u_\kappa$ and $\nu_0,\dots,\nu_{\kappa-1}$ only, so they lie in $E_\kappa$; hence so does $M_\kappa-\Pi_\kappa M_\kappa$ (recall $\Pi_\kappa$ is the orthogonal projection onto $\spn(w_0,\dots,w_{\kappa-1})$). The vector $M_\kappa-\Pi_\kappa M_\kappa$ is orthogonal to every $w_s$ with $s<\kappa$ by the definition of an orthogonal projection. This proves the first case since $g^{(\kappa)} \propto M_\kappa-\Pi_\kappa M_\kappa$, and $g^{(\kappa)}$ has norm one by construction.

\step{Step 2: The later atoms.} Let $j\ge\kappa$. The atom $w_j$ splits into a part in $E_\kappa$ and a part built from frame vectors outside $E_\kappa$:
\[
w_j=\rho_jM_\kappa+\Big(\rho_j\sum_{\kappa<i\le j}\sqrt{\delta_i}\,u_i+\sqrt{f_j}\,\nu_j\Big).
\]
Since $g^{(\kappa)}$ lies in $E_\kappa$, only the first part contributes:
\[
\ip{g^{(\kappa)}}{w_j}=\rho_j\ip{g^{(\kappa)}}{M_\kappa}=-\frac{\rho_j}{c_\kappa}\ip{M_\kappa-\Pi_\kappa M_\kappa}{M_\kappa}=-\rho_j\,c_\kappa .
\]
The last equality holds because $M_\kappa-\Pi_\kappa M_\kappa$ is orthogonal to $\Pi_\kappa M_\kappa$, so its inner product with $M_\kappa$ is its squared norm, whose value is $c_\kappa^2$.

\step{Step 3: The ball.} This is the Cauchy--Schwarz inequality, since $g^{(\kappa)}$ has norm one and $b$ has norm at most~$r$.
\end{proof}

In the proof we keep track of how much of the frame the learner could know. We say that the process is at \emph{level} $L$, for $0\le L\le N-1$, when the first $2L+1$ vectors of the frame are revealed:
\[
u_0,\ \nu_0,\ u_1,\ \nu_1,\ \dots,\ u_{L-1},\ \nu_{L-1},\ u_L .
\]
At level $L$ the atoms $w_0,\dots,w_{L-1}$ and the walk position $M_L$ are functions of the revealed vectors. We call these atoms \emph{revealed} and the other atoms \emph{unrevealed}. Let $U_L$ be the span of the unrevealed frame vectors,
\[
U_L=\spn(\nu_L,\ u_{L+1},\ \nu_{L+1},\ \dots,\ u_{N-1},\ \nu_{N-1}),
\]
and let $\Proj_{U_L}$ be the orthogonal projection onto it. Every unrevealed atom, $j\ge L$, splits into a known part and a hidden part:
\begin{equation}\label{eq:split}
w_j=\underbrace{\rho_j\,M_L}_{\text{known}}+\underbrace{a^{(L)}_j}_{\text{hidden}},\qquad
a^{(L)}_j=\rho_j\sum_{L<i\le j}\sqrt{\delta_i}\,u_i+\sqrt{f_j}\,\nu_j\ \in U_L .
\end{equation}
The known parts of all unrevealed atoms are multiples of the same vector $M_L$, and they differ only through the shrinkage factors. This is the precise sense in which unrevealed atoms look alike to the learner. The hidden part $a^{(L)}_j$ is the component of $w_j$ orthogonal to the revealed vectors, so its norm is at most $\norm{w_j}\le1$ (\cref{lem:sizes}). Finally, the loss vector $g^{(\kappa)}$ is a function of the vectors revealed at level $\kappa$.

\subsection{Four deterministic lemmas}\label{sec:lemmas}

This subsection proves four facts about the geometry of $K$. The body is small (\cref{lem:sizes}). The per-round costs~$c_\kappa$ are not too small (\cref{lem:level}). The hidden parts of unrevealed atoms cannot cancel each other (\cref{lem:cancel}). As a consequence, a point of $K$ with a small hidden component earns little (\cref{lem:earn}). No randomness is involved.

\begin{lemma}[Body]\label{lem:sizes}
Every point of $K$ has norm at most $1$, and the diameter of $K$ is at most $\sqrt5/2<1.12$.
\end{lemma}

\begin{proof}
\firststep{Step 1: Norms for all points of $K$.} The frame is orthonormal, so the squared norm of an atom is the sum of its squared coordinates:
\[
\norm{w_j}^2=\rho_j^2\,e_j+f_j\ \le\ e_j+f_j\ \le\ 1
\]
by \cref{lem:profile}(f). The ball has radius $r<1$ and the origin has norm $0$. A convex combination of points of norm at most one has norm at most one. Hence, every point of $K$ has norm at most $1$.

\step{Step 2: Distance between two atoms.} Let $i<j$. The two atoms $w_i, w_j$ share the steps $0,\dots,i$ of the walk (with different shrinkage factors) and differ by the steps $i+1,\dots,j$ and by their private noises. Reading off coordinates, we get
\[
\norm{w_i-w_j}^2=\underbrace{\rho_j^2(e_j-e_i)}_{\text{steps }i+1,\dots,j}+\underbrace{(\rho_i-\rho_j)^2e_i}_{\text{shared steps}}+\underbrace{f_i+f_j}_{\text{noises}} .
\]
The first and last terms add up to at most $e_j-e_i+f_i+f_j$, which is at most $1$ by \cref{lem:profile}(f). The middle term is at most $\frac14$, because the shrinkage factors lie in $[\frac12,1]$. So the distance is at most $\sqrt5/2$.

\step{Step 3: Distances for the other pairs.} An atom is at distance at most $1+r<\sqrt5/2$ from the origin and from every point of the ball, and two points of the ball are at distance at most $2r$. The diameter of a convex hull is the largest distance between two of its generating points.
Hence, we get $\diam(K) \le \sqrt{5}/2 < 1.12$.
\end{proof}

The next lemma lower bounds the per-round costs $c_\kappa$. Its proof reduces to the recursion of \cref{lem:qrec}.

\begin{lemma}[Lower bound on per-round costs]\label{lem:level}
For every $0\le\kappa\le N-1$, we have $c_\kappa\ge(4N)^{-1/4}$.
\end{lemma}

This lemma says that each loss vector rewards the unrevealed atoms by an amount of order $N^{-1/4}$; it is where the rate $T\cdot N^{-1/4}$ comes from.

\begin{proof}
By definition $c_\kappa^2=\norm{M_\kappa-\Pi_\kappa M_\kappa}^2=\min_{y_0,\dots,y_{\kappa-1}}\norm{M_\kappa-\sum_{s<\kappa}y_s\,w_s}^2$
(recall $\Pi_\kappa$ is the orthogonal projection onto $\spn(\{w_0,\dots,w_{\kappa-1}\})$).
The computation of Step~1 in the proof of \cref{lem:qrec}, with the shrunk atoms $w_s=\rho_sM_s+\sqrt{f_s}\,\nu_s$ and the cumulative quantities $Z_i=1-\sum_{s=i}^{\kappa-1}\rho_sy_s$, gives the coordinate $\sqrt{\delta_i}\,Z_i$ along~$u_i$ and $-y_s\sqrt{f_s}=(\sqrt{f_s}/\rho_s)(Z_s-Z_{s+1})$ along~$\nu_s$, hence
\[
c_\kappa^2=\min_{Z_0,\dots,Z_{\kappa-1}}\ \Big[\ \sum_{i=0}^{\kappa}\delta_i\,Z_i^2+\sum_{s=0}^{\kappa-1}\frac{f_s}{\rho_s^2}\,(Z_s-Z_{s+1})^2\ \Big]\qquad\text{with }Z_\kappa=1 .
\]
Since $\rho_s\le1$, the coefficients $f_s/\rho_s^2$ are at least $f_s$ for all $s$, so for every $Z_0,\dots,Z_{\kappa-1}$ the bracket is at least the one of \cref{lem:qrec}, and we get $c_\kappa^2\ge q_\kappa\ge1/(2\sqrt N)$, that is $c_\kappa \ge (4N)^{-1/4}$.
\end{proof}

A learner might try to use unrevealed atoms without knowing their hidden parts, by mixing them with a ball point that cancels those parts. The next lemma rules this out: a nonnegative combination of hidden parts of unrevealed atoms has norm proportional to its total weight. Since the ball has radius $r$, the total weight on unrevealed atoms in such a mixture is at most about $4N^{1/4}r=\frac1{10}$ (\cref{lem:earn}).

\begin{lemma}[Atom cancellation]\label{lem:cancel}
For every level $L$ and all nonnegative weights $\lambda_L,\dots,\lambda_{N-1}$ with sum $\Lambda$,
\[
\Big\|\sum_{j\ge L}\lambda_j\,a^{(L)}_j\Big\|\ \ge\ \frac{\Lambda}{4N^{1/4}} .
\]
\end{lemma}

\begin{proof}
The proof runs the recursion of \cref{lem:qrec} from the tail of the chain, with the same parallel sum~\eqref{eq:par}.

\step{Step 1: Lower bound by a quadratic problem $r_L$.} 
Let $Y_i=\sum_{j\ge i}\lambda_j$ be the weight on the atoms $i,\dots,N-1$. Then $Y_L=\Lambda$, $Y_N=0$ and $\lambda_j=Y_j-Y_{j+1}$. By \eqref{eq:split}, the coordinate of $\sum_j\lambda_ja^{(L)}_j$ along $u_i$ (for $i>L$) is $\sqrt{\delta_i}\sum_{j\ge i}\rho_j\lambda_j$, and its coordinate along $\nu_j$ is $\sqrt{f_j}\,\lambda_j$.
The weights $\lambda_L,\dots,\lambda_{N-1}$ are nonnegative and $\rho_j\ge\frac12$, so the coordinate along $u_i$ is at least $\sqrt{\delta_i}\,Y_i/2$. Hence, we have
\begin{equation}\label{eq:lb-quad-cancel}
\Big\|\sum_j\lambda_j\,a^{(L)}_j\Big\|^2\ \ge\ \frac14\Big[\ \sum_{i=L+1}^{N-1}\delta_i\,Y_i^2+\sum_{j=L}^{N-1}f_j\,(Y_j-Y_{j+1})^2\ \Big]
\end{equation}
(putting the factor $\frac14$ on the coordinates $\nu_j$, $L\le j\le N-1$, as well only weakens the bound). The bracket is a quadratic function of $Y_{L},\dots,Y_{N-1}$. Let $r_L$ be its minimum when $Y_L=1$ and $Y_N=0$. That is, for every $L<N$,
\[
r_L:=\min_{Y_{L+1},\dots,Y_{N-1}}\ 
\Big[\ \sum_{i=L+1}^{N-1}\delta_i\,Y_i^2+\sum_{j=L}^{N-1}f_j\,(Y_j-Y_{j+1})^2\ \Big]
\qquad\text{with }Y_L=1 \text{ and } Y_N=0.
\]
By homogeneity the bracket in~\eqref{eq:lb-quad-cancel} is at least $\Lambda^2\,r_L$. Thus, we are left to prove that $r_L \ge 1/(4\sqrt N)$.

\step{Step 2: The sequence $(r_L)_{L<N}$ satisfies a recursion formula.} For $L=N-1$ there is no free variable and $r_{N-1}=f_{N-1}$. For $L<N-1$, the problem defining $r_L$ consists of the problem for $r_{L+1}$, with $Y_{L+1}$ now free, plus the two terms $\delta_{L+1}Y_{L+1}^2$ and $f_L(1-Y_{L+1})^2$. If $Y_{L+1}$ is fixed to a value $t$ instead of $1$ in the definition of $r_{L+1}$, the minimum becomes $r_{L+1}\,t^2$ by homogeneity. Thus, minimizing first over $Y_{L+2},\dots,Y_{N-1}$ and then over $Y_{L+1}$ in the definition of $r_L$ gives
\begin{align*}
r_L
&=\min_{Y_{L+1}}\Big[\ f_L\,(1-Y_{L+1})^2+\delta_{L+1}Y_{L+1}^2+\min_{Y_{L+2},\dots,Y_{N-1}}\Big[\ \sum_{i=L+2}^{N-1}\delta_i\,Y_i^2+\sum_{j=L+1}^{N-1}f_j\,(Y_j-Y_{j+1})^2\ \Big]\ \Big]\\
&=\min_{Y_{L+1}}\Big[\ f_L\,(1-Y_{L+1})^2+\big(\delta_{L+1}+r_{L+1}\big)\,Y_{L+1}^2\ \Big],
\end{align*}
and using \eqref{eq:par} gives the recursion
\[
r_{N-1}=f_{N-1},\qquad r_L=f_L\psum\big(\delta_{L+1}+r_{L+1}\big)\quad(L<N-1).
\]

\step{Step 3: The recursion stays above $1/(4\sqrt N)$.} Let $R_L=\sqrt N\,r_L$. We prove by descending induction that $R_L\ge 1/4$ for all $L<N$. At $L=N-1$, we have $R_{N-1}=\sqrt{N}f_{N-1}=h_0/2=\frac12$ by \cref{lem:profile}(b), since $b_{N-1}=0$. Suppose $R_{L+1}\ge 1/4$, and let $a=a_{L+1}\ge1$ and $b=b_L$. By \cref{lem:profile}(b,c) and the monotonicity of the parallel sum,
\[
R_L\ \ge\ \frac{h_b}{2}\psum\Big(\frac1{\psi_a}+\frac14\Big).
\]
Using $1/(x\psum y)=1/x+1/y$, the right side is at least $\frac14$ exactly when
\[
\frac{2}{h_b}+\frac{4\psi_a}{4+\psi_a}\ \le\ 4,\qquad\text{that is,}\qquad 4+\psi_a\ \le\ 8h_b .
\]
By \cref{lem:profile}(d,e) we have $h_b\ge h_{a-1}\ge1$ and $\psi_a=h_a+h_{a-1}\le2h_{a-1}+1$. Thus we have
\[
4+\psi_a\ \le\ 5+2h_{a-1}\ \le\ 8h_{a-1}\ \le\ 8h_b ,
\]
and $R_L\ge\frac14$. Hence, we get $r_L\ge1/(4\sqrt N)$ for all $L<N$.

\step{Step 4: Conclusion.} Steps 1 to 3 give that the squared norm is at least $\Lambda^2/(16\sqrt N)$, which is the claim.
\end{proof}

Apart from the ball, only unrevealed atoms earn, that is, have a negative loss (\cref{lem:F1}), and putting weight on those atoms requires a hidden component (\cref{lem:cancel}). The next lemma combines these two facts.

\begin{lemma}[What a point of $K$ can earn]\label{lem:earn}
Let $L$ be a level, let $L\le\kappa\le N-1$, and let $x\in K$. Then
\[
\ip{g^{(\kappa)}}{x}\ \ge\ -\,4N^{1/4}\,c_\kappa\Big(\norm{\Proj_{U_L}x}+r\Big)\ -\ r .
\]
\end{lemma}

The bound should be compared with the loss $-c_\kappa/2$ of the comparator. If the hidden component $\norm{\Proj_{U_L}x}$ is negligible, the right side is about $-\frac1{10}c_\kappa-r$, because $4N^{1/4}r=\frac1{10}$.

\begin{proof}
\firststep{Step 1: Write $x$ as a mixture.} Every point of $K$ is a convex combination of the origin, the atoms and one point of the ball:
\[
x=\sum_j\lambda_j\,w_j+\mu\,b,\qquad \lambda_j,\mu\ge0,\qquad \sum_j\lambda_j+\mu\le1,\qquad b\in B_r .
\]

\step{Step 2: The hidden component controls the weight on unrevealed atoms.} Revealed atoms lie in the span of the revealed vectors, so they have no component in $U_L$, while by \eqref{eq:split} each unrevealed atom contributes its hidden part. Projecting onto $U_L$, we get
\[
\Proj_{U_L}x=\sum_{j\ge L}\lambda_j\,a^{(L)}_j+\mu\,\Proj_{U_L}b .
\]
The last term has norm at most $r$, so the combination $\sum_{j\ge L}\lambda_j\,a^{(L)}_j$ of hidden parts has norm at most $\norm{\Proj_{U_L}x}+r$. \Cref{lem:cancel} then bounds the weight $\Lambda=\sum_{j\ge L}\lambda_j$ on unrevealed atoms:
\[
\Lambda\ \le\ 4N^{1/4}\Big(\norm{\Proj_{U_L}x}+r\Big).
\]

\step{Step 3: Compute the loss.} By \cref{lem:F1}, atoms before $\kappa$ contribute nothing, each atom from $\kappa$ on contributes $-\rho_jc_\kappa\ge-c_\kappa$ per unit of weight, and the ball point contributes at least $-r$. Since $\kappa\ge L$, the atoms from $\kappa$ on are unrevealed, so their total weight is at most $\Lambda$. Hence $\ip{g^{(\kappa)}}x\ge-c_\kappa\Lambda-r$, and Step 2 gives the claim.
\end{proof}

\Cref{lem:earn} applies to every point of $K$. We never assume that the learner plays combinations of oracle answers: it may play any point it can certify to lie in $K$, for instance using its knowledge of the distribution of the instance.

\subsection{The clean process}\label{sec:clean}

We want to show that each oracle call reveals at most the next atom. The difficulty is circular: the oracle's answers depend on the unrevealed frame, so the learner's later messages depend on it too, and then we cannot use the randomness of the unrevealed frame to control those messages. To break the circle we introduce a \emph{clean oracle} that answers using only the revealed vectors. The \emph{clean process} is the run of (a copy of) the learner against this oracle, as in \cref{sec:all-outline}. \Cref{lem:agree} shows that it agrees with the true oracle as long as the learner's messages are nearly orthogonal to the unrevealed frame, and \cref{sec:hidden} shows that this holds with high probability. The clean process is an analysis device: the actual run is against the true oracle, and \cref{sec:proof} shows that the two runs coincide with high probability.

\begin{definition}[Clean oracle]\label{def:clean}
Fix a margin parameter $\tau$ with $0<\tau\le\gamma r/5$. The clean oracle keeps a level~$L$, initially~$0$, and handles queries and losses as follows.

\emph{Queries.} For the query $d=0$ it answers $0$. For a query $d\ne0$ at level~$L$, it computes three numbers from the revealed vectors:
\[
V=\min\Big(0,\ \min_{s<L}\ip d{w_s},\ -r\norm d\Big),\qquad
A=\rho_L\ip d{M_L},\qquad
\beta=\tau\norm d .
\]
Here $V$ is the best value among the candidates known at level $L$, namely $0$, the revealed atoms $w_0,\dots,w_{L-1}$ (none when $L=0$) and the ball point $-r\,d/\norm d$, and $p_V$ denotes the first of them, in this order (the order of the oracle's list in \cref{sec:all-instance}), that attains $V$. The number $A$ is the value the next atom $w_L$ would have if its hidden part were zero, and $\beta$ is a margin for the effect of hidden parts. The clean oracle's answer has two cases:
\begin{itemize}[leftmargin=*,itemsep=2pt]
\item[(C1)] If $A\ge V+\beta$, the answer is $p_V$ and the level does not change. The reason: even allowing for the hidden parts, no unrevealed atom can beat the known candidates.
\item[(C2)] Otherwise, the level is raised to $L+1$, which reveals $\nu_L$ and $u_{L+1}$ and hence the atom $w_L$. The answer is the first candidate among $0,w_0,\dots,w_L$ and $-r\,d/\norm d$ that minimizes $\ip d\cdot$. The reason: the next atom might win, so we reveal it and compare exactly.
\end{itemize}

\emph{Losses.} After the play of round $t$, the loss $g_t=g^{(\kappa_t)}$ is announced, and the level is raised to $\kappa_t$ if it is lower. This is needed because the loss vector is a function of the vectors revealed at level $\kappa_t$.
\end{definition}

Every answer of the clean oracle, and every change of its level, is computed from the learner's messages and the revealed vectors only. Rule (C2) requires $L\le N-2$; the next lemma shows that this always holds.

\begin{lemma}[Level counting]\label{lem:count}
In the clean process, the $i$-th query of round $t\ge1$ ($1\le i\le m$) is made at level at most $m(t-1)+i-1$, which is at most $\kappa_t-1\le N-2$. The play of round $t$ is made at level at most $\kappa_t=mt$.
\end{lemma}

\begin{proof}
Each query raises the level by at most one, by construction. Round $0$ makes no query, so $x_0$ is played at level $0=\kappa_0$, and the loss $g_0=g^{(0)}$ leaves the level at $0$. Suppose that the level is at most $\kappa_{t-1}=m(t-1)$ when round $t\ge1$ starts. Then the $i$-th query of the round is made at level at most $m(t-1)+i-1$, and the play at level at most $m(t-1)+m=\kappa_t$. When the loss is announced the level becomes exactly $\kappa_t=mt$, which is the induction hypothesis for round $t+1$. Finally $\kappa_t-1\le m(T-1)-1=N-2$.
\end{proof}

The next lemma shows that clean oracle answers are true oracle answers for queries with small hidden component.

\begin{lemma}\label{lem:agree}
Let $d$ be a query made at level $L\le N-2$, and suppose that $d=0$ or $\norm{\Proj_{U_L}d}<\tau\norm d$. Then the true oracle's answer $\LMO(d)$ is the clean oracle answer.
\end{lemma}

\begin{proof}
If $d=0$ both answers are $0$, so let $d\ne0$. Recall the definitions of $V, A, \beta$ in \Cref{def:clean}.

\step{Step 1: The values of the unrevealed atoms.} By \eqref{eq:split}, every unrevealed atom has value
\[
\ip d{w_j}=\rho_j\,A_0+\eps_j,\qquad\text{where}\qquad A_0=\ip d{M_L}\quad\text{and}\quad \eps_j=\ip{\Proj_{U_L}d}{a^{(L)}_j}.
\]
The known quantity $A_0$ is common to all unrevealed atoms. The hidden term $\eps_j$ is small: by the assumption $\norm{\Proj_{U_L}d}<\tau\norm d$, and since $\norm{a^{(L)}_j}\le1$, we have $|\eps_j|<\beta$. The ball point has value $-r\norm d$, so $V\le-r\norm d$, and since $\tau\le\gamma r/5\le r/10$ this is well below $-\beta$:
\[
V\ \le\ -r\norm d\ <\ -\beta .
\]
The true oracle returns the first candidate of the full list $0,w_0,\dots,w_{N-1},-r\,d/\norm d$ that attains the true minimum. We show that the unrevealed atoms never change the clean answer.

\step{Step 2: Under rule (C1) no unrevealed atom reaches $V$.} Suppose $A\ge V+\beta$, so the clean oracle uses rule (C1). If $A_0\ge0$, every unrevealed atom has value greater than $-\beta$, hence greater than $V$. If $A_0<0$, the shrinkage makes later atoms worse, not better: $\rho_j\le\rho_L$ gives $\rho_jA_0\ge\rho_LA_0=A$, so the value is greater than $A-\beta\ge V$. In both cases every unrevealed atom is strictly above $V$. So the true minimum is $V$, and the first candidate attaining it is $p_V$.

\step{Step 3: Under rule (C2) no atom beyond $w_L$ can win.} Otherwise, $A< V+\beta$ and the clean oracle uses rule~(C2). Suppose an atom $w_j$ with $j>L$ has value at most $V$ (otherwise, the true oracle answers is the clean oracle answers under rule (C2)). We show that $w_L$ is then strictly better, so $w_j$ is not a minimizer.

First, the known part of the query must point strongly towards the walk. Indeed $\rho_jA_0+\eps_j\le V\le-r\norm d$ and $\eps_j>-\beta$, so
\[
\rho_j\,A_0\ <\ -r\norm d+\beta\ <\ 0,\qquad\text{hence}\qquad |A_0|\ >\ r\norm d-\beta\ \ge\ \tfrac9{10}\,r\norm d,
\]
using $\rho_j\le1$ and $\beta\le r\norm d/10$. This is where the ball matters: since it offers the value $-r\norm d$, an unrevealed atom can compete only if $|A_0|$ is of order $r\norm d$.

Second, the shrinkage turns this into a strict preference for $w_L$. Since $A_0<0$ and $\rho_L-\rho_j\ge\gamma$,
\[
\ip d{w_L}-\ip d{w_j}=-(\rho_L-\rho_j)\,|A_0|+\eps_L-\eps_j\ <\ -\tfrac9{10}\,\gamma r\norm d+2\tau\norm d\ \le\ -\tfrac12\,\gamma r\norm d\ <\ 0,
\]
using $\tau\le\gamma r/5$. So the minimizers of $\ip d\cdot$ in the full list are among $0,w_0,\dots,w_L$ and the ball point. These appear in the same order in the full list and in the list used by (C2), so the true answer is the clean answer.
\end{proof}

The proof shows how the two design choices work together. Because the ball offers $-r\norm d$ for free, an unrevealed atom can win only if $|A_0|$ is at least about $r\norm d$. When $|A_0|$ is that large, the shrinkage gives the next atom an advantage of at least $\gamma|A_0|\approx\gamma r\norm d$, which exceeds any perturbation from the hidden parts, at most $2\tau\norm d$.

\subsection{Hidden directions stay hidden with high probability}\label{sec:hidden}

It remains to show that the hypothesis of \cref{lem:agree} holds for every query with high probability, and that the plays have small hidden components, as needed in \cref{lem:earn}. Let $v_1,v_2,\dots,v_Q$ be all messages of the learner in the clean process (its queries and its plays, in order), so that $Q\le m(T-1)+T\le2N$. Let $L_i$ be the level when $v_i$ is sent, $\ell_i=2L_i+1$ the number of revealed frame vectors, and $U_i=U_{L_i}$. In this section, we denote by $\phi_1,\dots,\phi_{2N}$ the random orthonormal vectors $u_0,\nu_0,\dots,u_{N-1},\nu_{N-1}$ in the same order.

The argument has two parts. First, in the clean process every message is a function of the revealed vectors (\cref{lem:prefix}). Second, given the revealed vectors, the rest of the frame is uniformly random, and a fixed vector has a tiny projection onto a random low-dimensional subspace of a high-dimensional space (\cref{lem:proj}).

\begin{lemma}[Prefix property]\label{lem:prefix}
For every $i$ and $\ell$, the event $\{\ell_i=\ell\}$, and the message $v_i$ on this event, are functions of $\phi_1,\dots,\phi_\ell$.
\end{lemma}

\begin{proof}
The clean process uses the frame only through revealed vectors, and it reveals them one at a time in the fixed order $\phi_1,\phi_2,\dots$. The decision to reveal the next vector depends only on the vectors already revealed and on the learner's messages. The learner is deterministic, so its messages are functions of the answers and losses it has received, which are themselves computed from revealed vectors. By induction over the steps of the process, everything that happens before the $(\ell+1)$-th vector is revealed is a function of $\phi_1,\dots,\phi_\ell$. Both claims follow.
\end{proof}

\begin{lemma}[Projection onto a random subspace]\label{lem:proj}
Let $W$ be a Euclidean space of dimension $n_W$, let $U$ be a uniformly random $k$-dimensional subspace of $W$, and let $e\in W$ be a fixed unit vector. For every $s\ge0$ with $\sqrt{n_W}>1+s$,
\[
\Prob\Big[\ \norm{\Proj_Ue}\ \ge\ \frac{\sqrt k+s}{\sqrt{n_W}-1-s}\ \Big]\ \le\ 2e^{-s^2/2}.
\]
\end{lemma}

The bound says that $\norm{\Proj_Ue}$ is at most about $\sqrt{k/n_W}$ with high probability: a fixed direction has a tiny projection onto a random subspace of much smaller dimension.

\begin{proof}
\firststep{Step 1: Reduce to a Gaussian vector.} By rotation invariance we may instead fix $U$, say as the span of the first $k$ coordinates of $W$, and take $e$ uniformly distributed on the unit sphere of $W$. A uniform point of the sphere can be written $z/\norm z$ with $z$ a standard Gaussian vector of $W$, so $\norm{\Proj_Ue}$ has the law of $\norm{z_{1:k}}/\norm z$, where $z_{1:k}$ denotes the first $k$ coordinates.

\step{Step 2: The numerator and the denominator concentrate.} The maps $z\mapsto\norm{z_{1:k}}$ and $z\mapsto\norm z$ are $1$-Lipschitz, so by Gaussian concentration~\citep[Thm.~5.6]{BLM13} each is within $s$ of its mean, on the relevant side, with probability at least $1-e^{-s^2/2}$. The means satisfy $\E\norm{z_{1:k}}\le\sqrt k$, by Jensen's inequality, and $\E\norm z\ge\sqrt{n_W-1}\ge\sqrt{n_W}-1$, because the variance of $\norm z$ is at most one by the Gaussian Poincar\'e inequality. On the intersection of the two events, the ratio is below the stated threshold.
\end{proof}

\begin{proposition}[The good event]\label{prop:good}
Let $\delta\in(0,1)$ and $s=\sqrt{2\log(2Q/\delta)}$, assume $\sqrt{n-2N}>1+s$, and let
\[
\tau_n=\frac{\sqrt{2N}+s}{\sqrt{n-2N}-1-s}.
\]
If $\tau_n\le\tau$, then with probability at least $1-\delta$ every message satisfies $v_i=0$ or $\norm{\Proj_{U_i}v_i}<\tau\norm{v_i}$. We call this event $\mathcal E$.
\end{proposition}

\begin{proof}
\firststep{Step 1: Condition on the revealed vectors.} Fix a message index $i$ and a prefix length $\ell$, and condition on $\phi_1,\dots,\phi_\ell$. By \cref{lem:prefix}, on the event $\{\ell_i=\ell\}$ the message $v_i$ is then a fixed vector.

\step{Step 2: The rest of the frame is still random.} Because the frame is uniformly random, its remaining vectors $\phi_{\ell+1},\dots,\phi_{2N}$ form, conditionally, a uniformly random orthonormal family in the orthogonal complement $W$ of the revealed vectors. So $U_i$ is a uniformly random subspace of $W$, of dimension $k=2N-\ell\le2N$, while $W$ has dimension $n_W=n-\ell\ge n-2N$.

\step{Step 3: Apply \cref{lem:proj}.} There is nothing to prove if $v_i=0$, or if $v_i$ is orthogonal to $W$ (then $\Proj_{U_i}v_i=0$). Otherwise let $e$ be the normalized projection of $v_i$ onto $W$. Since $U_i\subset W$, the projection of $v_i$ onto $U_i$ is its projection onto $W$ followed by $\Proj_{U_i}$, so $\norm{\Proj_{U_i}v_i}\le\norm{v_i}\,\norm{\Proj_{U_i}e}$. The threshold of \cref{lem:proj} with these $k$ and $n_W$ is at most $\tau_n$, so the conditional probability that $\norm{\Proj_{U_i}v_i}\ge\tau_n\norm{v_i}$ is at most $2e^{-s^2/2}=\delta/Q$.

\step{Step 4: Remove the conditioning.} Averaging over $\phi_1,\dots,\phi_\ell$ and summing over $\ell$, message $i$ fails with probability at most $\delta/Q$. A union bound over the at most $Q$ messages, together with $\tau_n\le\tau$, gives the claim.
\end{proof}

\subsection{Putting the pieces together}\label{sec:proof}

\step{Step 1: parameters.} We take
\[
r=\tfrac1{40}N^{-1/4},\qquad \tau=\frac{\gamma r}5=\frac{N^{-1/4}}{400(N-1)},\qquad \delta=\tfrac1{100}N^{-1/4},\qquad n\ge1.1\cdot10^6\,N^{7/2}.
\]
Each choice has a reason.
\begin{itemize}[leftmargin=*,itemsep=1pt]
\item The ball radius $r$ is small enough that the ball earns the learner at most $r=\frac1{40}N^{-1/4}$ per round and lets it fake a weight of at most $4N^{1/4}r=\frac1{10}$ on unrevealed atoms (\cref{lem:earn}).
\item The margin $\tau$ is the largest allowed by \cref{lem:agree}.
\item The failure probability $\delta$ is small enough that failures cost little (Step 5).
\item The dimension $n$ makes a random direction's projection onto the unrevealed frame, of order $\sqrt{N/n}$, smaller than $\tau\approx N^{-5/4}$. Precisely, with $Q\le2N$ we have $2Q/\delta\le400N^{5/4}$, and a direct computation shows that $n\ge1.1\cdot10^6N^{7/2}$ implies $\sqrt{n-2N}>1+s$ and $\tau_n\le\tau$ for every $N\ge2$. For large $N$ it suffices that $n\ge3.2\cdot10^5N^{7/2}$.
\end{itemize}
By \cref{prop:good}, the good event $\mathcal E$ has probability at least $1-\delta$.

\step{Step 2: on the good event the learner is in the clean process.} Run the learner against the true oracle and, in parallel, against the clean oracle. On $\mathcal E$ the two runs produce the same messages and answers. We check this message by message. As long as all answers so far agree, the next message is the same in both runs, because the learner is deterministic and the losses are fixed. If this message is a query, it is made at a level at most $N-2$ (\cref{lem:count}) and, on $\mathcal E$, it satisfies the hypothesis of \cref{lem:agree}; so the two answers agree again.

\step{Step 3: what the learner knows at each play.} By \cref{lem:count}, the play $x_t$ of round $t$ is made at a level $L_t\le\kappa_t$. So every atom with index at least $\kappa_t$, the comparator included, is unrevealed at that time. On $\mathcal E$ the play also has a small hidden component, $\norm{\Proj_{U_{L_t}}x_t}\le\tau\norm{x_t}\le\tau$, using $\norm{x_t}\le1$ (\cref{lem:sizes}).

\step{Step 4: the regret of one round.} \Cref{lem:earn}, applied with $\kappa=\kappa_t$ and $L=L_t$, bounds what the play earns:
\[
\ip{g_t}{x_t}\ \ge\ -c_{\kappa_t}\cdot4N^{1/4}(\tau+r)-r\ \ge\ -0.11\,c_{\kappa_t}-r,
\]
since $4N^{1/4}r=0.1$ and $4N^{1/4}\tau=1/(100(N-1))\le0.01$. The comparator earns $c_{\kappa_t}/2$ by \cref{lem:F1}, because its index $N-1$ is at least $\kappa_t$. With \cref{lem:level} and the value of $r$,
\[
\ip{g_t}{x_t-x^*}\ \ge\ 0.39\,c_{\kappa_t}-r\ \ge\ \Big(\frac{0.39}{\sqrt2}-\frac1{40}\Big)N^{-1/4}\ \ge\ 0.25\,N^{-1/4}.
\]
Summing over the $T$ rounds, the regret against $x^*$, and hence the regret, is at least $0.25\,T\,N^{-1/4}$ on $\mathcal E$.

\step{Step 5: the expectation.} Off the good event we use the trivial bound: the regret is at least $-T$ times the diameter, which is at least $-1.12\,T$ by \cref{lem:sizes}. Therefore
\[
\E[\Reg_T]\ \ge\ 0.25\,TN^{-1/4}\,(1-\delta)-1.12\,T\delta\ \ge\ (0.25-0.0025-0.0112)\,TN^{-1/4}\ \ge\ 0.23\,TN^{-1/4}.
\]
This proves \cref{thm:all}.\hfill$\square$

\section{Proofs for Section~\ref{sec:ub}: the upper bound}\label{app:ub}

We use the notation of \cref{sec:ub}: rounds are numbered $t=1,\dots,T$, $G_t=\sum_{s\le t}g_s$, and $\Reg_T=\sum_{t=1}^T\ip{g_t}{x_t-x_\star}$ for a fixed comparator $x_\star\in K$. For convex losses $\ell_t$ with $g_t\in\partial\ell_t(x_t)$, convexity gives $\sum_t\ell_t(x_t)-\ell_t(x_\star)\le \Reg_T$, so it suffices to bound $\Reg_T$, and we work with the linear losses $x\mapsto\ip{g_t}x$ throughout. We only use $\norm{g_t}\le L$ and $\norm{p-q}\le D$ for all $p,q\in K$, and the fact that the minimizer $x_*$ of a differentiable convex function $f$ over $K$ satisfies $\ip{\nabla f(x_*)}{x_*-x}\le0$ for every $x\in K$.

The proofs follow those of \OPT for $m=1$ step by step. The only new ingredient is that a round of \cref{alg:mofw-opt} performs $m$ Frank--Wolfe steps instead of one, so that the contraction factor $1-2\sigma$ of a single step becomes $(1-2\sigma)^m=1-u$, with $u$ as in \eqref{eq:u}; for $m=1$, $u=2\sigma$ and every statement below is the corresponding statement of \OPT.

\paragraph{The exact minimizers and the potential.} Let $y_1:=x_1$ and, for $t\ge1$,
\[
 y_{t+1}:=\argmin_{y\in K}\ \eta\,\ip{G_t}{y}+\tfrac12\norm{y-x_1}^2 ,
\]
the iterates of follow-the-regularized-leader run on the same gradients. The potential is
\[
 \phi_t:=\sum_{s=1}^{t}\ip{g_s}{x_s-y_{t+1}}-\frac{1}{2\eta}\norm{y_{t+1}-x_1}^2+\frac{1}{6\eta}\norm{x_{t+1}-y_{t+1}}^2,\qquad t=0,\dots,T,
\]
so that $\phi_0=0$. Its first two terms are the potential of the classical regret analysis of follow-the-regularized-leader; the third one records how far the Frank--Wolfe iterate is from the exact minimizer.

\subsection{Fixed parameters}
\label{sec:proof:ofw-fixed}

\begin{lemma}[Per-round inequality]\label{lem:pot}
Let $\sigma\in(0,\tfrac12]$, $\eta>0$, and $u$, $\lambda^g=8\eta/(3u)$ as in \eqref{eq:u}. For every $t\in\{1,\dots,T\}$,
\[
 \phi_t-\phi_{t-1}\ \le\ \lambda^g\,\norm{g_t}^2+\frac{\sigma^2}{6\eta}\sum_{k=1}^{m}(1-2\sigma)^{m-k}\,\norm{z_{t,k-1}-v_{t,k}}^2 .
\]
For $m=1$ the sum is $\norm{x_t-v_t}^2$ and $\lambda^g=4\eta/(3\sigma)$, which is the lemma of \OPT.
\end{lemma}

\begin{proof}
Fix $t$ and write $P:=x_t-y_{t+1}$ and $Y:=y_t-y_{t+1}$, so that $x_t-y_t=P-Y$.

\firststep{Step 1: the regularizer terms.} By definition of the potential,
\begin{multline*}
 \phi_t-\phi_{t-1}=\ip{g_t}{P}+\ip{G_{t-1}}{Y}\\
 +\frac1{6\eta}\bigl(\norm{x_{t+1}-y_{t+1}}^2-\norm{x_t-y_t}^2\bigr)+\frac1{2\eta}\bigl(\norm{y_t-x_1}^2-\norm{y_{t+1}-x_1}^2\bigr).
\end{multline*}
The optimality of $y_t$ for $\eta\ip{G_{t-1}}{\cdot}+\frac12\norm{\cdot-x_1}^2$ gives $\ip{G_{t-1}}{Y}\le-\frac1\eta\ip{y_t-x_1}{Y}$, and $-2\ip{y_t-x_1}{Y}+\norm{y_t-x_1}^2-\norm{y_{t+1}-x_1}^2=-\norm{Y}^2$. Hence
\begin{equation}\label{eq:pot1}
 \phi_t-\phi_{t-1}\ \le\ \ip{g_t}{P}-\frac1{2\eta}\norm{Y}^2+\frac1{6\eta}\bigl(\norm{x_{t+1}-y_{t+1}}^2-\norm{P-Y}^2\bigr).
\end{equation}

\step{Step 2: $m$ Frank--Wolfe steps contract the distance to $y_{t+1}$.} Fix $k\in\{1,\dots,m\}$. The optimality of $v_{t,k}$ for the linear function $\ip{\eta G_t+z_{t,k-1}-x_1}{\cdot}$ and that of $y_{t+1}$ give
\[
 \ip{G_t}{v_{t,k}-y_{t+1}}+\frac1\eta\ip{z_{t,k-1}-x_1}{v_{t,k}-y_{t+1}}\le0,\qquad
 \ip{G_t}{y_{t+1}-v_{t,k}}+\frac1\eta\ip{y_{t+1}-x_1}{y_{t+1}-v_{t,k}}\le0 ,
\]
and adding them, $\ip{z_{t,k-1}-y_{t+1}}{v_{t,k}-y_{t+1}}\le0$. Since $z_{t,k}=(1-\sigma)z_{t,k-1}+\sigma v_{t,k}$,
\begin{align*}
 \norm{z_{t,k}-y_{t+1}}^2&=(1-\sigma)^2\norm{z_{t,k-1}-y_{t+1}}^2+2\sigma(1-\sigma)\ip{z_{t,k-1}-y_{t+1}}{v_{t,k}-y_{t+1}}+\sigma^2\norm{v_{t,k}-y_{t+1}}^2\\
 &\le(1-\sigma)^2\norm{z_{t,k-1}-y_{t+1}}^2-2\sigma^2\ip{z_{t,k-1}-y_{t+1}}{v_{t,k}-y_{t+1}}+\sigma^2\norm{v_{t,k}-y_{t+1}}^2\\
 &=(1-2\sigma)\norm{z_{t,k-1}-y_{t+1}}^2+\sigma^2\norm{z_{t,k-1}-v_{t,k}}^2 ,
\end{align*}
where the inequality multiplies the nonpositive inner product by $-2\sigma^2$ instead of the larger coefficient $2\sigma(1-\sigma)$. For $m=1$ this is the computation of \OPT. Iterating it from $z_{t,0}=x_t$ to $z_{t,m}=x_{t+1}$, which uses $1-2\sigma\ge0$,
\begin{equation}\label{eq:pot-iter}
 \norm{x_{t+1}-y_{t+1}}^2\ \le\ (1-u)\norm{P}^2+\sigma^2\sum_{k=1}^m(1-2\sigma)^{m-k}\norm{z_{t,k-1}-v_{t,k}}^2 .
\end{equation}
Inserting \eqref{eq:pot-iter} into \eqref{eq:pot1},
\begin{equation}\label{eq:pot2}
 \phi_t-\phi_{t-1}\ \le\ \frac{\sigma^2}{6\eta}\sum_{k=1}^m(1-2\sigma)^{m-k}\norm{z_{t,k-1}-v_{t,k}}^2
 +\ip{g_t}{P}-\frac1{2\eta}\norm{Y}^2+\frac{1}{6\eta}\bigl((1-u)\norm{P}^2-\norm{P-Y}^2\bigr).
\end{equation}

\step{Step 3: trading the gradient term for a quadratic form.} Let $\lambda\ge0$. The optimality of $y_t$ and of $y_{t+1}$ give $\ip{G_{t-1}}{Y}+\frac1\eta\ip{y_t-x_1}{Y}\le0$ and $-\ip{G_t}{Y}-\frac1\eta\ip{y_{t+1}-x_1}{Y}\le0$; adding them, $-\ip{g_t}{Y}\le-\frac1\eta\norm{Y}^2$. Hence, by Young's inequality $\ip ab\le\lambda^g\norm a^2+\norm b^2/(4\lambda^g)$,
\[
 \ip{g_t}{P}=\ip{g_t}{P+\lambda Y}-\lambda\ip{g_t}{Y}\ \le\ \lambda^g\norm{g_t}^2+\frac{1}{4\lambda^g}\norm{P+\lambda Y}^2-\frac\lambda\eta\norm{Y}^2 .
\]
With \eqref{eq:pot2} this gives
\[
 \phi_t-\phi_{t-1}\ \le\ \lambda^g\norm{g_t}^2+\frac{\sigma^2}{6\eta}\sum_{k=1}^m(1-2\sigma)^{m-k}\norm{z_{t,k-1}-v_{t,k}}^2-Q(\lambda),
\]
\[
 Q(\lambda):=\frac{1+2\lambda}{2\eta}\norm{Y}^2-\frac1{4\lambda^g}\norm{P+\lambda Y}^2-\frac1{6\eta}\bigl((1-u)\norm{P}^2-\norm{P-Y}^2\bigr),
\]
and it remains to find $\lambda\ge0$ with $Q(\lambda)\ge0$.

\step{Step 4: $Q(\lambda)$ is a sum of squares.} Expanding the squares,
\[
 Q(\lambda)=\Bigl(\frac{2+3\lambda}{3\eta}-\frac{\lambda^2}{4\lambda^g}\Bigr)\norm{Y}^2+\Bigl(\frac{u}{6\eta}-\frac1{4\lambda^g}\Bigr)\norm{P}^2-2\Bigl(\frac1{6\eta}+\frac{\lambda}{4\lambda^g}\Bigr)\ip{P}{Y}.
\]
Take $\lambda=4/(3u)$. With $\lambda^g=8\eta/(3u)$, that is $1/(4\lambda^g)=3u/(32\eta)$, the three coefficients are $\frac{4u+7}{6u\eta}$, $\frac{7u}{96\eta}$ and $\frac{7}{24\eta}$, and
\[
 Q(\lambda)=\frac{7}{96\,\eta u}\,\norm{uP-4Y}^2+\frac{2}{3\eta}\norm{Y}^2\ \ge\ 0 ,
\]
since $\frac{7}{96\eta u}\norm{uP-4Y}^2=\frac{7u}{96\eta}\norm P^2-\frac{7}{12\eta}\ip PY+\frac{7}{6\eta u}\norm Y^2$ and $\frac{4u+7}{6u\eta}-\frac{7}{6u\eta}=\frac{2}{3\eta}$. This proves the lemma. For $m=1$, \OPT show by a Schur-complement argument that $Q(\lambda)\ge0$ for a whole interval of values of $\lambda$; the value $\lambda=4/(3u)$ belongs to it and makes the sum of squares explicit.
\end{proof}

\begin{theorem}[Potential-function bound]\label{thm:pot}
Let $\sigma\in(0,\tfrac12]$, $\eta>0$ and $u,\lambda^g$ as in \eqref{eq:u}. For every $x_\star\in K$,
\[
 \Reg_T\ \le\ \lambda^g\sum_{t=1}^T\norm{g_t}^2+\frac{\sigma^2}{6\eta}\sum_{t=1}^T\sum_{k=1}^m(1-2\sigma)^{m-k}\norm{z_{t,k-1}-v_{t,k}}^2+\frac{1}{2\eta}\norm{x_\star-x_1}^2
 \ \le\ \frac{8\eta L^2T}{3u}+\frac{\sigma u\,D^2T}{12\eta}+\frac{D^2}{2\eta} .
\]
\end{theorem}

\begin{proof}
Summing \cref{lem:pot} over $t=1,\dots,T$ and using $\phi_0=0$ bounds $\phi_T$ by the first two sums. Since $y_{T+1}$ minimizes $\eta\ip{G_T}{\cdot}+\frac12\norm{\cdot-x_1}^2$ over $K$, its value at $y_{T+1}$ is at most its value at $x_\star$, which reads
\[
 \sum_{t=1}^T\ip{g_t}{x_t-x_\star}-\frac{\norm{x_\star-x_1}^2}{2\eta}\ \le\ \sum_{t=1}^T\ip{g_t}{x_t-y_{T+1}}-\frac{\norm{y_{T+1}-x_1}^2}{2\eta}\ \le\ \phi_T ,
\]
the last step dropping $\frac1{6\eta}\norm{x_{T+1}-y_{T+1}}^2\ge0$. This is the first bound. For the second, $\norm{g_t}\le L$, $\norm{z_{t,k-1}-v_{t,k}}\le D$, $\norm{x_\star-x_1}\le D$, and $\sigma^2\sum_{k=1}^m(1-2\sigma)^{m-k}=\sigma^2\,\frac{1-(1-2\sigma)^m}{2\sigma}=\frac{\sigma u}{2}$.
\end{proof}

The tuning of \cref{cor:pot} balances the three terms of \cref{thm:pot}. Two elementary facts about $u=1-(1-2\sigma)^m$ are used: by Bernoulli's inequality $(1-2\sigma)^m\ge1-2m\sigma$, so
\begin{equation}\label{eq:u-upper}
 u\ \le\ 2m\sigma ,
\end{equation}
and $(1-2\sigma)^m\le e^{-2m\sigma}$ together with $e^{\eps}\ge1+\eps$ gives, for $\eps:=2m\sigma$,
\begin{equation}\label{eq:u-lower}
 u\ \ge\ 1-e^{-\eps}\ \ge\ \frac{\eps}{1+\eps},\qquad\text{that is}\qquad \frac1u\ \le\ \frac1\eps+1 .
\end{equation}
For $m=1$ both are equalities, $u=2\sigma=\eps$.

\begin{proof}[Proof of \cref{cor:pot}]
Here $\eta=\frac{D}{2L}(3/T)^{3/4}m^{1/4}$ and $\sigma=\sqrt{3/(mT)}$, which is at most $\frac12$ because $mT\ge12$; note $m\sigma^2T=3$ and $\eta L^2T=\frac{LD}{2}\,3^{3/4}(mT)^{1/4}$. We bound the three terms of \cref{thm:pot}.
\emph{The regularizer term:} $\frac{D^2}{2\eta}=\frac{LD}{3^{3/4}}\,T^{3/4}m^{-1/4}$.
\emph{The atom term:} by \eqref{eq:u-upper}, $\frac{\sigma uD^2T}{12\eta}\le\frac{m\sigma^2D^2T}{6\eta}=\frac{D^2}{2\eta}$.
\emph{The gradient term:} by \eqref{eq:u-lower} with $\eps=2m\sigma=2\sqrt{3m/T}$,
\[
 \frac{8\eta L^2T}{3u}\ \le\ \frac{8\eta L^2T}{3\eps}+\frac{8\eta L^2T}{3}
 =\frac{4\eta L^2T}{3m\sigma}+\frac{8\eta L^2T}{3}
 =\frac{2LD}{3^{3/4}}\,T^{3/4}m^{-1/4}+\frac{4LD}{3^{1/4}}\,(mT)^{1/4}.
\]
Adding the three bounds gives the claim. For $m=1$ the two facts are equalities, so the atom term equals $D^2/(2\eta)$ and the gradient term equals $\frac{2LD}{3^{3/4}}T^{3/4}$: this is the bound of \OPT.
\end{proof}

\subsection{Anytime parameters}\label{sub:pot-anytime}

Let $(\eta_t,\sigma_t)_{t\ge2}$ be parameters with $\eta_t$ non-increasing and $\sigma_t\in(0,\tfrac12]$. As in \OPT, round $t$ uses the parameters of index $t+1$:
\[
 \dir_{t,k}=\eta_{t+1}G_t+(z_{t,k-1}-x_1),\qquad z_{t,k}=(1-\sigma_{t+1})z_{t,k-1}+\sigma_{t+1}v_{t,k},\qquad
 y_{t+1}=\argmin_{y\in K}\ \eta_{t+1}\ip{G_t}{y}+\tfrac12\norm{y-x_1}^2 ,
\]
with $y_1=x_1$, and the potential is
\[
 \phi_t:=\sum_{s=1}^{t}\ip{g_s}{x_s-y_{t+1}}-\frac{1}{2\eta_{t+1}}\norm{y_{t+1}-x_1}^2+\frac{1}{6\eta_{t+1}}\norm{x_{t+1}-y_{t+1}}^2 .
\]
For $t\ge3$ let
\[
 u_t:=1-(1-2\sigma_t)^m,\qquad d_t:=1-\frac{\eta_t}{\eta_{t-1}}\in[0,1),\qquad \delta_t:=\frac{d_t}{u_t},\qquad
 \lambda^g_t:=\frac{8\eta_t}{3u_t}\cdot\frac{1}{1-\frac32\delta_t}\quad\bigl(\text{when }\tfrac32\delta_t<1\bigr).
\]
The number $d_t$ measures the growth of the regularization weight from one round to the next, $\frac1{\eta_t}-\frac1{\eta_{t-1}}=\frac{d_t}{\eta_t}$. With constant parameters $d_t=0$ and $\lambda^g_t$ is the constant of \cref{lem:pot}; the growing weight costs the factor $(1-\frac32\delta_t)^{-1}$ on the gradient term and nothing else.

\begin{lemma}[Per-round inequality, time-varying parameters]\label{lem:pot-any}
For every $t\ge2$ with $\tfrac32\delta_{t+1}<1$,
\[
 \phi_t-\phi_{t-1}\ \le\ \lambda^g_{t+1}\norm{g_t}^2+\frac{\sigma_{t+1}^2}{6\eta_{t+1}}\sum_{k=1}^m(1-2\sigma_{t+1})^{m-k}\norm{z_{t,k-1}-v_{t,k}}^2 .
\]
\end{lemma}

\begin{proof}
Fix $t\ge2$ and abbreviate $u:=u_{t+1}$, $\sigma:=\sigma_{t+1}$, $\delta:=\delta_{t+1}$, $\lambda:=4/(3u)$ and $\lambda^g:=\lambda^g_{t+1}$. Let $\Delta:=\frac1{\eta_{t+1}}-\frac1{\eta_t}=\frac{\delta u}{\eta_{t+1}}\ge0$, and write $P:=x_t-y_{t+1}$, $Y:=y_t-y_{t+1}$ and $r:=y_{t+1}-x_1$, so that $x_t-y_t=P-Y$ and $y_t-x_1=Y+r$. We follow the proof of \cref{lem:pot}; the difference between $\eta_t$ and $\eta_{t+1}$ creates the extra terms carrying $\Delta$.

\firststep{Step 1.} Now
\[
 \phi_t-\phi_{t-1}=\ip{g_t}{P}+\ip{G_{t-1}}{Y}+\frac{\norm{x_{t+1}-y_{t+1}}^2}{6\eta_{t+1}}-\frac{\norm{P-Y}^2}{6\eta_t}+\frac{\norm{Y+r}^2}{2\eta_t}-\frac{\norm{r}^2}{2\eta_{t+1}} .
\]
The optimality of $y_t$ (weight $\eta_t$) gives $\ip{G_{t-1}}{Y}\le-\frac1{\eta_t}\ip{Y+r}{Y}$, and $\frac{\norm{Y+r}^2}{2\eta_t}-\frac{\ip{Y+r}{Y}}{\eta_t}=\frac{\norm r^2-\norm Y^2}{2\eta_t}$. Hence
\begin{equation}\label{eq:pot1-any}
 \phi_t-\phi_{t-1}\ \le\ \ip{g_t}{P}-\frac{\norm{Y}^2}{2\eta_t}-\frac\Delta2\norm{r}^2+\frac{\norm{x_{t+1}-y_{t+1}}^2}{6\eta_{t+1}}-\frac{\norm{P-Y}^2}{6\eta_t} ;
\end{equation}
the growing weight produces the free nonpositive term $-\frac\Delta2\norm r^2$.

\step{Step 2.} Step~2 of \cref{lem:pot} applies verbatim with $(\eta,\sigma)=(\eta_{t+1},\sigma_{t+1})$, since both optimality relations used there carry the weight $\eta_{t+1}$:
$\norm{x_{t+1}-y_{t+1}}^2\le(1-u)\norm{P}^2+\sigma^2\sum_k(1-2\sigma)^{m-k}\norm{z_{t,k-1}-v_{t,k}}^2$.

\step{Step 3.} The optimality of $y_t$ (weight $\eta_t$) and of $y_{t+1}$ (weight $\eta_{t+1}$) give $\ip{G_{t-1}}{Y}+\frac1{\eta_t}\ip{Y+r}{Y}\le0$ and $-\ip{G_t}{Y}-\frac1{\eta_{t+1}}\ip{r}{Y}\le0$; adding them, $-\ip{g_t}{Y}\le-\frac1{\eta_t}\norm{Y}^2+\Delta\ip{r}{Y}$. Young's inequality then gives
\[
 \ip{g_t}{P}=\ip{g_t}{P+\lambda Y}-\lambda\ip{g_t}{Y}\le\lambda^g\norm{g_t}^2+\frac{\norm{P+\lambda Y}^2}{4\lambda^g}-\frac{\lambda}{\eta_t}\norm{Y}^2+\lambda\Delta\ip{r}{Y}.
\]
The drift term is absorbed by the free term of Step~1, as in \OPT: $-\frac\Delta2\norm{r}^2+\lambda\Delta\ip{r}{Y}=-\frac\Delta2\norm{r-\lambda Y}^2+\frac{\lambda^2\Delta}2\norm{Y}^2\le\frac{\lambda^2\Delta}{2}\norm{Y}^2$. Combining with Steps~1 and~2,
\[
 \phi_t-\phi_{t-1}\le\lambda^g\norm{g_t}^2+\frac{\sigma^2}{6\eta_{t+1}}\sum_{k}(1-2\sigma)^{m-k}\norm{z_{t,k-1}-v_{t,k}}^2-\tilde Q(\lambda),
\]
\[
 \tilde Q(\lambda):=\Bigl(\frac{1+2\lambda}{2\eta_t}-\frac{\lambda^2\Delta}{2}\Bigr)\norm{Y}^2-\frac{\norm{P+\lambda Y}^2}{4\lambda^g}-\frac{1-u}{6\eta_{t+1}}\norm{P}^2+\frac{\norm{P-Y}^2}{6\eta_t} ,
\]
and it remains to show that $\tilde Q(\lambda)\ge0$.

\step{Step 4: comparison with the constant-parameter form.} For a symmetric $2\times2$ matrix $M$ write $q_M:=M_{11}\norm{Y}^2+2M_{12}\ip{Y}{P}+M_{22}\norm{P}^2$; if $M$ is positive semidefinite then $q_M\ge0$. Substituting $\frac1{\eta_t}=\frac1{\eta_{t+1}}-\Delta$ and expanding, $\tilde Q(\lambda)=q_N-\frac{\norm{P+\lambda Y}^2}{4\lambda^g}$ with
\[
 N=\frac{1}{\eta_{t+1}}\tilde N-\Delta E,\qquad
 \tilde N=\begin{pmatrix}\tfrac23+\lambda&-\tfrac16\\[2pt]-\tfrac16&\tfrac u6\end{pmatrix},\qquad
 E=\begin{pmatrix}\tfrac23+\lambda+\tfrac{\lambda^2}2&-\tfrac16\\[2pt]-\tfrac16&\tfrac16\end{pmatrix}.
\]
For constant parameters ($\Delta=0$) and $\lambda^g=8\eta_{t+1}/(3u)$, the expression $q_{\tilde N}/\eta_{t+1}-\norm{P+\lambda Y}^2/(4\lambda^g)$ is exactly the form $Q(\lambda)$ of Step~4 of \cref{lem:pot}, which is a sum of squares. We compare the two forms. Since $\Delta=\delta u/\eta_{t+1}$,
\[
 N-\Bigl(1-\frac32\delta\Bigr)\frac{\tilde N}{\eta_{t+1}}=\frac{\delta u}{\eta_{t+1}}\Bigl(\frac{3}{2u}\tilde N-E\Bigr),
 \qquad
 \frac{3}{2u}\tilde N-E=\begin{pmatrix}\dfrac{10-3u-6u^2}{9u^2}&\dfrac{2u-3}{12u}\\[8pt]\dfrac{2u-3}{12u}&\dfrac1{12}\end{pmatrix}
\]
for $\lambda=4/(3u)$, and the last matrix is positive semidefinite: its lower-right entry is positive and its determinant $\frac{13+24u-36u^2}{432u^2}$ is positive for $0<u\le1$, the numerator being concave with values $13$ and $1$ at $u=0$ and $u=1$. Hence $q_N\ge(1-\frac32\delta)\,q_{\tilde N}/\eta_{t+1}$. Finally $\lambda^g=\lambda^g_{t+1}$ was defined so that $(1-\frac32\delta)/(4\lambda^g)=1/(4\cdot8\eta_{t+1}/(3u))$, and therefore
\[
 \tilde Q(\lambda)\ \ge\ \Bigl(1-\frac32\delta\Bigr)\Bigl[\frac{q_{\tilde N}}{\eta_{t+1}}-\frac{3u}{32\eta_{t+1}}\norm{P+\lambda Y}^2\Bigr]=\Bigl(1-\frac32\delta\Bigr)Q(\lambda)\ \ge\ 0 ,
\]
where $Q(\lambda)$ is the form of \cref{lem:pot} for the parameters $(\eta_{t+1},\sigma_{t+1})$.
\end{proof}

\begin{theorem}[Anytime potential bound]\label{thm:pot-any}
Let $(\eta_t)_{t\ge2}$ be non-increasing, $\sigma_t\in(0,\tfrac12]$, and $\tfrac32\delta_t<1$ for $3\le t\le T+1$. Then for every $T\ge1$ and $x_\star\in K$,
\begin{align*}
 \Reg_T&\le\frac{\norm{x_\star-x_1}^2}{2\eta_{T+1}}+\phi_1+\sum_{t=2}^T\lambda^g_{t+1}\norm{g_t}^2
 +\sum_{t=2}^T\frac{\sigma_{t+1}^2}{6\eta_{t+1}}\sum_{k=1}^m(1-2\sigma_{t+1})^{m-k}\norm{z_{t,k-1}-v_{t,k}}^2\\
 &\le\frac{D^2}{2\eta_{T+1}}+LD+\frac{D^2}{6\eta_2}+L^2\sum_{t=2}^T\lambda^g_{t+1}+D^2\sum_{t=2}^T\frac{\sigma_{t+1}u_{t+1}}{12\eta_{t+1}} .
\end{align*}
\end{theorem}

\begin{proof}
As for \cref{thm:pot}: $\phi_T=\phi_1+\sum_{t=2}^T(\phi_t-\phi_{t-1})$, each increment is bounded by \cref{lem:pot-any}, and the optimality of $y_{T+1}$ (weight $\eta_{T+1}$) gives $\sum_{t\le T}\ip{g_t}{x_t-x_\star}-\frac{\norm{x_\star-x_1}^2}{2\eta_{T+1}}\le\phi_T$. For the second line, $\phi_1=\ip{g_1}{x_1-y_2}-\frac{\norm{y_2-x_1}^2}{2\eta_2}+\frac{\norm{x_2-y_2}^2}{6\eta_2}\le LD+\frac{D^2}{6\eta_2}$, and the other terms are bounded as in \cref{thm:pot}.
\end{proof}

\begin{proof}[Proof of \cref{cor:pot-any}]
Here $\eta_t=\frac{D}{2L}(3/t)^{3/4}m^{1/4}$ and $\sigma_t=\min\{\tfrac12,\sqrt{3/(mt)}\}$, so that $\eta_t L^2=\frac{LD}2\,3^{3/4}m^{1/4}t^{-3/4}$ and $\frac{D^2}{2t\eta_t}=\frac{LD}{3^{3/4}}\,m^{-1/4}t^{-1/4}$. Put $\bar\eps_t:=2\sqrt{3m/t}$, which equals $2m\sigma_t$ when $\sigma_t<\frac12$. We first collect three facts, valid for every $t\ge3$.
\begin{itemize}[leftmargin=2em,itemsep=1pt]
\item[(a)] $\sigma_t u_t\le6/t$: by \eqref{eq:u-upper} and $\sigma_t\le\sqrt{3/(mt)}$, $\sigma_t u_t\le2m\sigma_t^2\le6/t$.
\item[(b)] $1/u_t\le1+1/\bar\eps_t$: if $\sigma_t=\frac12$ then $u_t=1$; otherwise $\bar\eps_t=2m\sigma_t$ and this is \eqref{eq:u-lower}.
\item[(c)] $\tfrac32\delta_t\le\frac34$ and $\frac{1}{1-\frac32\delta_t}\le1+6\delta_t$: indeed $d_t=1-((t-1)/t)^{3/4}\le1/t$ because $x^{3/4}\ge x$ on $[0,1]$, so by (b) $\delta_t\le\frac1t\bigl(1+\frac1{\bar\eps_t}\bigr)=\frac1t+\frac{1}{2\sqrt{3mt}}\le\frac13+\frac16=\frac12$, and $\frac1{1-x}\le1+4x$ for $0\le x\le\frac34$.
\end{itemize}
By (c), \cref{thm:pot-any} applies. We bound its terms, with all sums over the parameter index $t\in\{3,\dots,T+1\}$ (the round being $t-1$) and the elementary estimates $\sum t^{-1/4}\le\int_2^{T+1}x^{-1/4}dx\le\frac43(T+1)^{3/4}$, $\sum t^{-3/4}\le4(T+1)^{1/4}$ and $\sum_{t\ge3}t^{-7/4}\le3^{-7/4}+\int_3^\infty x^{-7/4}dx\le0.74$.
\begin{itemize}[leftmargin=2em,itemsep=1pt]
\item The regularizer term is $\frac{D^2}{2\eta_{T+1}}=\frac{LD}{3^{3/4}}\,m^{-1/4}(T+1)^{3/4}$, and $LD+\frac{D^2}{6\eta_2}=LD\bigl(1+\frac13(\tfrac23)^{3/4}m^{-1/4}\bigr)\le1.25\,LD$.
\item By (a), the atom term of round $t-1$ is at most $\frac{D^2}{2t\eta_t}=\frac{LD}{3^{3/4}}m^{-1/4}t^{-1/4}$, and the sum is at most $\frac43\,\frac{LD}{3^{3/4}}\,m^{-1/4}(T+1)^{3/4}$.
\item By (b) and (c), $\lambda^g_t\le\frac{8\eta_t}{3u_t}+\frac{8\eta_t}{3u_t}\cdot6\delta_t\le\frac{8\eta_t}{3}\Bigl(\frac1{\bar\eps_t}+1\Bigr)+16\eta_t\frac{d_t}{u_t^2}$, and $\frac{d_t}{u_t^2}\le\frac1t\Bigl(1+\frac1{\bar\eps_t}\Bigr)^2\le\frac1t\Bigl(2+\frac{2}{\bar\eps_t^2}\Bigr)=\frac2t+\frac1{6m}$. The four resulting pieces of $L^2\sum\lambda^g_t$ are
\begin{gather*}
 \sum\frac{8\eta_t L^2}{3\bar\eps_t}=\frac{2LD}{3^{3/4}}m^{-1/4}\sum t^{-1/4}\le\frac83\,\frac{LD}{3^{3/4}}\,m^{-1/4}(T+1)^{3/4},\\
 \sum\frac{8\eta_t L^2}{3}=\frac{4LD}{3^{1/4}}m^{1/4}\sum t^{-3/4}\le\frac{16}{3^{1/4}}LD\,(m(T+1))^{1/4},\\
 \sum16\eta_t L^2\frac2t=16\cdot3^{3/4}LD\,m^{1/4}\sum t^{-7/4}\le27\,LD\,m^{1/4},\\
 \sum\frac{16\eta_t L^2}{6m}=\frac43\,3^{3/4}LD\,m^{-3/4}\sum t^{-3/4}\le12.2\,LD\,m^{-3/4}(T+1)^{1/4}.
\end{gather*}
\end{itemize}
The three leading terms add up to $(1+\frac43+\frac83)\,3^{-3/4}=5\cdot3^{-3/4}$ times $LD\,m^{-1/4}(T+1)^{3/4}$. For $T\ge2$ we have $(T+1)^{3/4}\le T^{3/4}+\frac34T^{-1/4}\le T^{3/4}+0.63$ and $(T+1)^{1/4}\le(\tfrac32)^{1/4}T^{1/4}\le1.11\,T^{1/4}$, while $m^{1/4}\le(mT)^{1/4}$ and $m^{-3/4}\le m^{1/4}$; hence the lower-order terms are at most $(1.25+5\cdot3^{-3/4}\cdot0.63+16\cdot3^{-1/4}\cdot1.11+27+12.2\cdot1.11)\,LD\,(mT)^{1/4}\le57\,LD\,(mT)^{1/4}$. For $T=1$ the claim is trivial since $\Reg_1\le LD$.
\end{proof}

\subsection{Proof for the capped tuning}\label{sec:proof-capped-tuning}

\begin{proof}[Proof of \cref{cor:pot-cap}, fixed-horizon part]
Now $\eta=\frac{D}{2L}(3/T)^{3/4}\kappa$ with $\kappa=\min\{m,T\}^{1/4}$, $\sigma=\sqrt{3/(mT)}\le\frac12$, and $\eta L^2T=\frac{LD}2\,3^{3/4}T^{1/4}\kappa$. As in the previous proof, $\frac{D^2}{2\eta}=\frac{LD}{3^{3/4}}\frac{T^{3/4}}{\kappa}$, the atom term is at most $\frac{D^2}{2\eta}$, and by \eqref{eq:u-lower}
\[
 \frac{8\eta L^2T}{3u}\ \le\ \frac{4\eta L^2T}{3m\sigma}+\frac{8\eta L^2T}{3}=\frac{2LD}{3^{3/4}}\,\frac{T^{3/4}\kappa}{\sqrt m}+\frac{4LD}{3^{1/4}}\,T^{1/4}\kappa .
\]
If $m\le T$, then $\kappa=m^{1/4}$, the three terms add up to $\frac{4LD}{3^{3/4}}T^{3/4}m^{-1/4}+\frac{4LD}{3^{1/4}}(mT)^{1/4}$, and $(mT)^{1/4}\le T^{3/4}m^{-1/4}$. If $m\ge T$, then $\kappa=T^{1/4}$, $T^{3/4}\kappa/\sqrt m=T/\sqrt m\le\sqrt T$ and $T^{1/4}\kappa=\sqrt T$, so the three terms add up to at most $\bigl(\frac{4}{3^{3/4}}+\frac{4}{3^{1/4}}\bigr)LD\sqrt T$. In both cases $\Reg_T\le\bigl(4\cdot3^{-3/4}+4\cdot3^{-1/4}\bigr)LD\,T^{3/4}\min\{m,T\}^{-1/4}$, and $4\cdot3^{-3/4}+4\cdot3^{-1/4}<4.8$.
\end{proof}

\begin{proof}[Proof of \cref{cor:pot-cap}, anytime part]
Write $\kappa_t:=\min\{m,t\}^{1/4}$, so that $\tilde\eta_t=\frac{D}{2L}(3/t)^{3/4}\kappa_t$, and $\Lambda:=\max\{(T+1)^{3/4}m^{-1/4},\sqrt{T+1}\}$. Since $\kappa_t$ is non-decreasing, $\tilde\eta_t$ is non-increasing and $\tilde\eta_t/\tilde\eta_{t-1}\ge((t-1)/t)^{3/4}$, so the facts (a), (b), (c) of the proof of \cref{cor:pot-any} hold, and \cref{thm:pot-any} gives, as in that proof,
\begin{equation}\label{eq:capped-start}
 \Reg_T\ \le\ \frac{D^2}{2\tilde\eta_{T+1}}+1.25\,LD+\sum_{t=3}^{T+1}\Bigl[\frac{D^2}{2t\tilde\eta_t}+\frac{8\tilde\eta_tL^2}{3}\Bigl(\frac1{\bar\eps_t}+1\Bigr)+16\,\tilde\eta_tL^2\Bigl(\frac2t+\frac1{6m}\Bigr)\Bigr],
\end{equation}
with $\bar\eps_t=2\sqrt{3m/t}$. We use $\tilde\eta_tL^2=\frac{LD}{2}\,3^{3/4}\,t^{-3/4}\kappa_t$ and $\frac{D^2}{2t\tilde\eta_t}=\frac{LD}{3^{3/4}}\,\frac{t^{-1/4}}{\kappa_t}$, the three elementary bounds
\[
 \kappa_t\le m^{1/4},\qquad \kappa_t\le t^{1/4},\qquad \frac1{\kappa_t}\le m^{-1/4}+t^{-1/4},
\]
and the integral estimates $\sum_{t=3}^{T+1}t^{-1/4}\le\frac43(T+1)^{3/4}$, $\sum_{t=3}^{T+1}t^{-1/2}\le2\sqrt{T+1}$ and $\sum_{t\ge3}t^{-3/2}\le1.35$. Both $(T+1)^{3/4}m^{-1/4}$ and $\sqrt{T+1}$ are at most $\Lambda$. The terms of \eqref{eq:capped-start} are bounded one by one:
\begin{align*}
 \frac{D^2}{2\tilde\eta_{T+1}}&=\frac{LD}{3^{3/4}}\,\frac{(T+1)^{3/4}}{\kappa_{T+1}}\le\frac{LD}{3^{3/4}}\bigl((T+1)^{3/4}m^{-1/4}+\sqrt{T+1}\bigr)\le0.9\,LD\,\Lambda,\\
 \sum_{t=3}^{T+1}\frac{D^2}{2t\tilde\eta_t}&\le\frac{LD}{3^{3/4}}\Bigl(m^{-1/4}\sum_{t=3}^{T+1}t^{-1/4}+\sum_{t=3}^{T+1}t^{-1/2}\Bigr)\le\frac{LD}{3^{3/4}}\Bigl(\frac43+2\Bigr)\Lambda\le1.5\,LD\,\Lambda,\\
 \sum_{t=3}^{T+1}\frac{8\tilde\eta_tL^2}{3\bar\eps_t}&=\frac{2\cdot3^{1/4}}{3}\,LD\sum_{t=3}^{T+1}\frac{\kappa_t\,t^{-1/4}}{\sqrt m}\le\frac{2\cdot3^{1/4}}{3}\,LD\,m^{-1/4}\sum_{t=3}^{T+1}t^{-1/4}\le1.2\,LD\,\Lambda,\\
 \sum_{t=3}^{T+1}\frac{8\tilde\eta_tL^2}{3}&=\frac43\,3^{3/4}LD\sum_{t=3}^{T+1}t^{-3/4}\kappa_t\le\frac43\,3^{3/4}LD\sum_{t=3}^{T+1}t^{-1/2}\le6.1\,LD\,\Lambda,\\
 \sum_{t=3}^{T+1}\frac{32\,\tilde\eta_tL^2}{t}&=16\cdot3^{3/4}LD\sum_{t=3}^{T+1}t^{-7/4}\kappa_t\le16\cdot3^{3/4}LD\sum_{t=3}^{T+1}t^{-3/2}\le50\,LD,\\
 \sum_{t=3}^{T+1}\frac{16\,\tilde\eta_tL^2}{6m}&\le\frac43\,3^{3/4}LD\sum_{t=3}^{T+1}t^{-3/4}\kappa_t\le6.1\,LD\,\Lambda .
\end{align*}
The first two lines use $1/\kappa_t\le m^{-1/4}+t^{-1/4}$, the third $\kappa_t\le m^{1/4}$, the last three $\kappa_t\le t^{1/4}$ and $m\ge1$. Adding up those bounds, we get $\Reg_T\le16\,LD\,\Lambda+52\,LD$. For $T\ge2$ we have $\Lambda\le(3/2)^{3/4}\max\{T^{3/4}m^{-1/4},\sqrt T\}\le1.4\max\{T^{3/4}m^{-1/4},\sqrt T\}$ and $\max\{T^{3/4}m^{-1/4},\sqrt T\}\ge\sqrt2$, so $\Reg_T\le(16\cdot1.4+52/\sqrt2)\,LD\max\{T^{3/4}m^{-1/4},\sqrt T\}\le60\,LD\max\{T^{3/4}m^{-1/4},\sqrt T\}$. For $T=1$ the claim is trivial since $\Reg_1\le LD$.
\end{proof}

\end{document}